\documentclass{article}

\usepackage{arxiv}

\usepackage[utf8]{inputenc}
\usepackage[T1]{fontenc}
\usepackage{hyperref}
\usepackage{url}
\usepackage{booktabs}
\usepackage{amsfonts}
\usepackage{nicefrac}
\usepackage[expansion=false]{microtype}
\usepackage{graphicx}
\usepackage{float}
\usepackage{amsmath,amssymb,amsfonts,amsthm}
\usepackage{multirow}
\usepackage{array}
\usepackage{xcolor}
\usepackage{caption}
\usepackage{enumitem}
\usepackage{subcaption}
\usepackage{cleveref}
\usepackage{setspace}
\usepackage{natbib}
\usepackage{doi}
\usepackage{placeins}
\usepackage{adjustbox}
\usepackage{pifont}
\usepackage{algorithm}
\usepackage{algorithmicx}
\usepackage{algpseudocode}

\title{Deep RL-Tuned Model-Free Adaptive Control for Lower-Limb Exoskeletons During Sit-to-Stand Transitions}

\author{
  Ranjeet~Kumbhar \\
  Department of Computer Science and Engineering\\
  Thapar Institute of Engineering and Technology\\
  Patiala, India -- 147004 \\
  \texttt{rkumbhar\_me24@thapar.edu}
  \And
  Appaso~M.~Gadade$^{*}$ \\
  Department of Mechanical Engineering\\
  Thapar Institute of Engineering and Technology\\
  Patiala, India -- 147004 \\
  \texttt{appaso.gadade@thapar.edu}
  \And
  Rajmeet~Singh \\
  Mechatronics Engineer, Research Office\\
  Mohamed bin Zayed University of Artificial Intelligence (MBZUAI)\\
  Abu Dhabi, UAE \\
  \texttt{rajmeet.bhourji@mbzuai.ac.ae}
  \And
  Ashish~Singla \\
  Department of Mechanical Engineering\\
  Thapar Institute of Engineering and Technology\\
  Patiala, India -- 147004 \\
  \texttt{ashish.singla@thapar.edu}
  \And
  Ravinder~Kumar \\
  Department of Computer Science and Engineering\\
  Thapar Institute of Engineering and Technology\\
  Patiala, India -- 147004 \\
  \texttt{ravinder@thapar.edu}
  \And
  \vspace{-1em}
  \normalsize\textit{$^{*}$ Corresponding author.}
}

\hypersetup{
  pdftitle={Deep RL-Tuned Model-Free Adaptive Control for Lower-Limb Exoskeletons During Sit-to-Stand Transitions},
  pdfsubject={Robotics, Reinforcement Learning, Assistive Robots, Model-Free Control},
  pdfauthor={Ranjeet Kumbhar, Appaso M. Gadade, Rajmeet Singh, Ashish Singla, Ravinder Kumar},
  pdfkeywords={Lower limb exoskeleton, Sit-to-stand, Reinforcement learning, Model-free control, Assistive robots},
}

\begin{document}
\maketitle

\begin{abstract}
Sit-to-stand (STS) transitions impose significant joint-loading demands on elderly individuals, making them a primary target for lower-limb exoskeleton assistance. However, accurate trajectory tracking during STS is challenging due to complex, time-varying human--exoskeleton interaction dynamics and inter-subject variability that render model-based control approaches difficult to apply in practice. This paper presents an intelligent model-free adaptive backstepping control strategy for a bilateral lower-limb exoskeleton during STS motion. The proposed controller design uses an ultra-local second-order model to avoid explicit system identification, while a Gaussian radial basis function (RBF) neural network estimates the unknown lumped dynamics online. To further improve phase-aware tracking performance, a Twin Delayed Deep Deterministic Policy Gradient (TD3) reinforcement learning agent is integrated as a supervisory gain scheduler that adaptively adjusts controller gains across the distinct phases of STS motion. The proposed controller is evaluated through co-simulation in MATLAB/Simulink and Simscape Multibody using OpenSim-derived reference trajectories and benchmarked against state-of-the-art controllers. Results demonstrate that the proposed controller achieves the lowest average RMSE of $0.078^\circ$ across all joints, representing improvements of 60.2\%, 54.4\%, 48.7\%, and 42.6\% over proportional integral derivative (PID), model-free adaptive control (MFAC), linear quadratic regulator (LQR), and sliding-mode control (SMC), respectively. TD3 integration further reduces tracking error by 35\%, 33\%, and 79\% at the hip, knee, and ankle joints compared to the standalone RBF-MFAC baseline. These results demonstrate the effectiveness and robustness of the proposed controller design for assistive exoskeleton control during STS transitions.
\end{abstract}


\section{Introduction}
\label{sec:introduction}

The global population is experiencing a significant demographic shift toward older age groups, mainly due to improved health care and higher standards of living. As a result, the proportion of elderly individuals continues to rise across both developed and developing nations. For example, older adults aged 65 and over constitute approximately 26\% of the population in Japan and around 18\% in the United Kingdom \citep{UN2017}. The demographic transition toward an older population is not confined to developed countries but is also increasingly visible in developing economies. In India, the share of the population classified as older adults is expected to be nearly 13\% of the total population in 2050 \citep{Shukla2020}. In addition, stroke is a severe cerebrovascular disease with sudden disruption of blood supply to the brain, often resulting in irreversible neurological damage, and it remains one of the leading causes of long-term disability worldwide \citep{Wang2022}. The global burden of stroke has increased substantially over recent decades, with incident cases rising by approximately 70\% and the number of people living with stroke increasing by 102\% between 1990 and 2019 \citep{Feigin2022}.

The sit-to-stand (STS) movement is one of the most physically challenging activities of daily living for older adults. Compared with walking, it requires coordinated whole-body motion, higher joint torque, and stability to achieve a successful postural transition \citep{Young2016}. Despite its complexity, STS is one of the most frequently performed functional tasks in daily life. A healthy adult executes approximately $60 \pm 22$ STS transitions per day \citep{Dall2009}. Furthermore, the ability to perform STS is considered a requirement for independent ambulation and functional mobility. STS performance is widely used in clinical practice as a measure of functional capability and lower-extremity muscle strength due to its strong correlation with lower-limb strength, balance, and mobility \citep{Alcazar2018}.

The increasing incidence of mobility impairments and the challenges of daily activities have highlighted the need for advanced lower-limb rehabilitation robots that can accurately understand and respond to patients' movement intentions, enabling safer and more efficient rehabilitation training as the aging population continues to grow and the demand for long-term care increases. Over the past decades, significant advances have been made in this field, leading to the development of several rehabilitation-oriented exoskeleton systems. For example, Ekso \citep{Pransky2014}, ALEX \citep{Banala2010}, and LOKOMAT \citep{Bernhardt2005} have been widely employed to assist patients with neurological impairments and mobility limitations during rehabilitation training.

The lower-limb exoskeletons exhibit highly nonlinear, strongly coupled dynamics due to the complex interactions between the human musculoskeletal system and the robot. As a result, developing effective control strategies is critical to providing coordinated, stable, and safe assistance during lower-limb movements \citep{Li2017}. Many studies on STS assistance have shown that powered knee and lower-limb exoskeletons can significantly reduce user effort while also providing additional joint extension support during movement. However, the effectiveness of such assistance depends heavily on precise torque control, balance aware motion assistance, phase-dependent control strategies, and adaptation to the user's remaining physical capabilities \citep{Aliman2017,Shepherd2017,Vantilt2019,Huo2021}. Furthermore, recent studies on optimal control and motion capture-based approaches for elderly users have shown that STS assistance remains a difficult problem, as maintaining balance while accommodating user-specific movement characteristics must be addressed concurrently \citep{Roelker2022}. In addition, predictive biomechanical models have been studied to evaluate the force transmission capabilities of passive exoskeletons, providing valuable insights for optimizing assistance effectiveness and human exoskeleton interaction \citep{Fernandez2026}. Consequently, significant research efforts have been devoted to the development of advanced control strategies for exoskeletons such as parameter identification methods based on triple-loop iterative frameworks combined with backpropagation neural networks \citep{Cheng2026}, robust nonsingular fast terminal sliding-mode control approaches \citep{Hernandez2020}, model-based control techniques \citep{Vantilt2019}, impedance modulation control methods \citep{Huo2021}, adaptive backstepping sliding mode control \citep{Narayan2024}, fuzzy logic \citep{Sharma2021}, adaptive SMC variants using fuzzy logic \citep{Yang2020}, and a variety of approaches ranging from phase-based assistance algorithms \citep{Liu2019} and sophisticated multisensory control frameworks \citep{Fleischer2008} to adaptive patient-cooperative control strategies \citep{Yu2026}.

Despite significant progress in exoskeleton technology, several control challenges remain unexplored. The human machine system is a highly nonlinear, time varying plant whose complexity increases with the number of controlled degrees of freedom (DOF), while practical operation introduces additional uncertainties in the form of external disturbances, measurement noise, and unmodeled interaction dynamics. Conventional feedback controllers, such as proportional integral derivative (PID), are simple to implement but perform poorly under large parameter variations or external perturbations. Model based approaches such as computed torque control (CTC), linear quadratic regulator (LQR), and sliding-mode control (SMC) provide improved robustness against such uncertainties. However, their practical applicability is often constrained by the need for accurate system models and computationally intensive gain-tuning procedures \citep{Vantilt2019}.

The limitations of model-dependent strategies have driven substantial interest in model-free and adaptive control methods that avoid the need for explicit system identification. To simplify control design in the presence of uncertainties and disturbances, model-free control based on the ultra-local model concept was introduced in \citep{Fliess2013}, where complex system dynamics are replaced by a simplified local representation. Unlike conventional model based methods, the ultra-local model primarily depends on the measurable input output signals and does not require accurate system parameters. Therefore, model free control has become a promising approach for exoskeleton applications due to its simplicity and adaptability \citep{Amara2025}.

While model free and neural adaptive controllers solve the system identification problem, their fixed-gain structures limit adaptability across dynamically distinct phases of complex transitions. Reinforcement learning (RL) has gained increasing attention as a complementary paradigm, owing to its ability to learn optimal policies through direct interaction with the environment without the need for explicit dynamic models. Deep RL extends this to continuous high dimensional action spaces via neural function approximation, following the study \citep{Khan2019} that demonstrated Q-learning and deep deterministic policy gradient (DDPG) for torque mapping.

Despite these advances, existing RL formulations have primarily been applied as standalone controllers rather than supervisory layers within analytically grounded adaptive frameworks. To the best of the authors' knowledge, no existing work has concurrently integrated a model-free adaptive backstepping controller, an online Gaussian radial basis function (RBF) neural estimator, and a Twin Delayed Deep Deterministic Policy Gradient (TD3) supervisory gain scheduler into a unified framework for STS trajectory tracking in lower-limb exoskeletons. The main contributions of this work are summarized as follows:

\begin{itemize}[leftmargin=*, itemsep=2pt]
\item An online Gaussian RBF neural estimator is integrated to approximate the unknown lumped dynamics in real time through an adaptive weight update law, providing stable disturbance estimation without requiring offline system identification.
\item A TD3 reinforcement learning agent is used as a supervisory gain scheduler to continuously adapt backstepping controller gains across the five phases of STS motion, resulting in phase aware tracking without manual retuning.
\item The proposed controller is validated through high-fidelity co-simulation integrating SolidWorks, Simscape Multibody, and OpenSim derived reference trajectories. It is benchmarked against several state-of-the-art controllers to evaluate its tracking performance, robustness, and control effectiveness.
\item The contribution of the TD3-based supervisory gain scheduler is quantitatively evaluated by comparing the standalone radial basis function-based model-free adaptive control (RBF-MFAC) controller with the proposed control design using root mean square error (RMSE), mean absolute error (MAE), and peak tracking error metrics, demonstrating substantial improvements in tracking performance across the hip, knee, and ankle joints.
\end{itemize}

\section{System Description}
\label{sec:system}

This section describes the human exoskeleton system developed for STS motion assistance. The lower-limb exoskeleton is a 10-DOF bilateral assistive device designed in the Computer-Aided Design (CAD) software (SolidWorks 2018) environment to structurally mirror human anthropometric distributions and imported into MATLAB-Simulink Simscape Multibody for dynamic simulation. The combined human exoskeleton system is modeled with a total mass of approximately 75~kg, equivalent to a human subject with the exoskeleton. The desired joint trajectories for the STS task are derived from OpenSim \citep{Delp2007} biomechanical kinematic data, providing physiologically accurate reference signals for hip, knee, and ankle joints. The controller is then applied to track these references and guide the exoskeleton through the complete STS motion.

\subsection{Mechanical Design of the Lower-Limb Exoskeleton}
\label{subsec:mechanical}

\begin{figure}[!ht]
\centering
\includegraphics[width=0.6\linewidth]{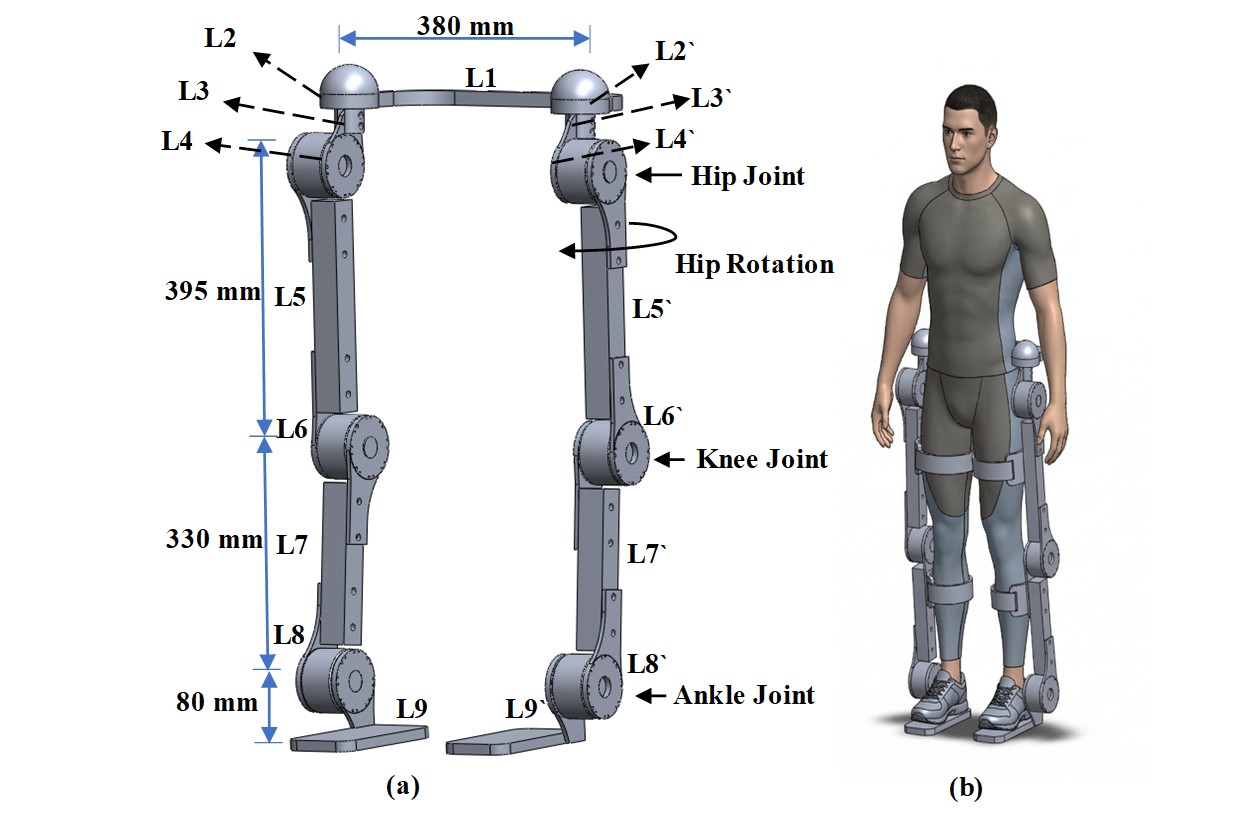}
\caption{Lower-limb exoskeleton mechanical structure: (a) CAD model illustrating link parameters and joint locations; (b) assembled human exoskeleton system \citep{Kumbhar2026}.}
\label{fig:fig1}
\end{figure}

The lower-limb exoskeleton is a bilateral assistive device that supports both legs simultaneously during the STS motion. The mechanical structure consists of a pelvis backboard, two thigh links, two shank links, and two foot-plate assemblies, connected through revolute joints at the hip, knee, and ankle of each leg. The exoskeleton has a total of eight active rotary joints, four on each leg, as shown in Fig.~\ref{fig:fig1}.

Each leg incorporates a hip rotation joint, a hip joint (flexion/extension), a knee joint (flexion/extension), and an ankle joint (plantarflexion/dorsiflexion). Accordingly, the hip, knee, and ankle joints are actively controlled, while the hip rotation joint remains mechanically present in the structure but is not actuated during STS simulation. This joint configuration ensures that the lower-limb exoskeleton can provide targeted assistance at the joints most mechanically loaded during rising from a seated position. Table~\ref{tab:link_params} summarizes the geometric and inertial properties of each link extracted from the CAD model.

\begin{table}[!ht]
\centering
\caption{Physical parameters of the lower-limb exoskeleton links.}
\label{tab:link_params}
\small
\begin{tabular}{@{}llcccc@{}}
\toprule
\textbf{Link} & \textbf{Segment} & \textbf{Length (mm)} & \textbf{Mass (kg)} & \textbf{Center of mass (m)} & \textbf{Moment of inertia (g\,m$^2$)} \\
\midrule
L1 & Backboard            & 380 & 0.3575 & $[0, 0.010, 0.086]$       & $[1.655, 10.700, 9.067]$ \\
L2 & Hip Rotation Bracket & --  & 0.1823 & $[0, 0.059, 0]$           & $[0.1245, 0.0964, 0.1257]$ \\
L3 & Hip Flex Connector   & --  & 0.0537 & $[0.003, 0.011, 0.003]$   & $[0.0801, 0.0289, 0.0517]$ \\
L4 & Hip Joint Connector  & --  & 0.0682 & $[0.003, 0.030, 0.007]$   & $[0.1851, 0.034, 0.1517]$ \\
L5 & Thigh Link           & 395 & 0.4594 & $[0, 0.1475, 0]$          & $[3.358, 0.1241, 3.357]$ \\
L6 & Knee Joint Connector & --  & 0.0682 & $[0.003, 0.030, 0.0067]$  & $[0.1851, 0.0340, 0.1517]$ \\
L7 & Shank Link           & 330 & 0.3474 & $[0, 0.1125, 0]$          & $[1.510, 0.0942, 1.508]$ \\
L8 & Ankle Joint Connector& --  & 0.0682 & $[0.003, 0.031, 0.0068]$  & $[0.1851, 0.034, 0.1517]$ \\
L9 & Foot                 & 80  & 0.3656 & $[0.046, 0.064, -0.05]$   & $[1.455, 1.624, 0.6283]$ \\
\bottomrule
\end{tabular}
\end{table}

The structural components of the exoskeleton were designed using 6061-T6 aluminum alloy, selected for its high strength-to-weight ratio, corrosion resistance, and machinability, making it well suited for wearable assistive devices. The material properties used in the simulation are listed in Table~\ref{tab:material}.

\begin{table}[!ht]
\centering
\caption{Material properties of 6061-T6 aluminum alloy.}
\label{tab:material}
\small
\begin{tabular}{@{}lcc@{}}
\toprule
\textbf{Property} & \textbf{Value} & \textbf{Unit} \\
\midrule
Young's Modulus               & $6.9 \times 10^{10}$ & Pa \\
Poisson's Ratio               & 0.33                & -- \\
Density                       & 2700                & kg/m$^3$ \\
Yield Strength                & $2.75 \times 10^{8}$ & Pa \\
Ultimate Tensile Strength     & $3.10 \times 10^{8}$ & Pa \\
Thermal Expansion Coefficient & $2.4 \times 10^{-5}$ & 1/K \\
\bottomrule
\end{tabular}
\end{table}

\subsection{STS Motion and Reference Trajectories}
\label{subsec:sts_motion}

\begin{figure}[!ht]
\centering
\includegraphics[width=0.85\linewidth]{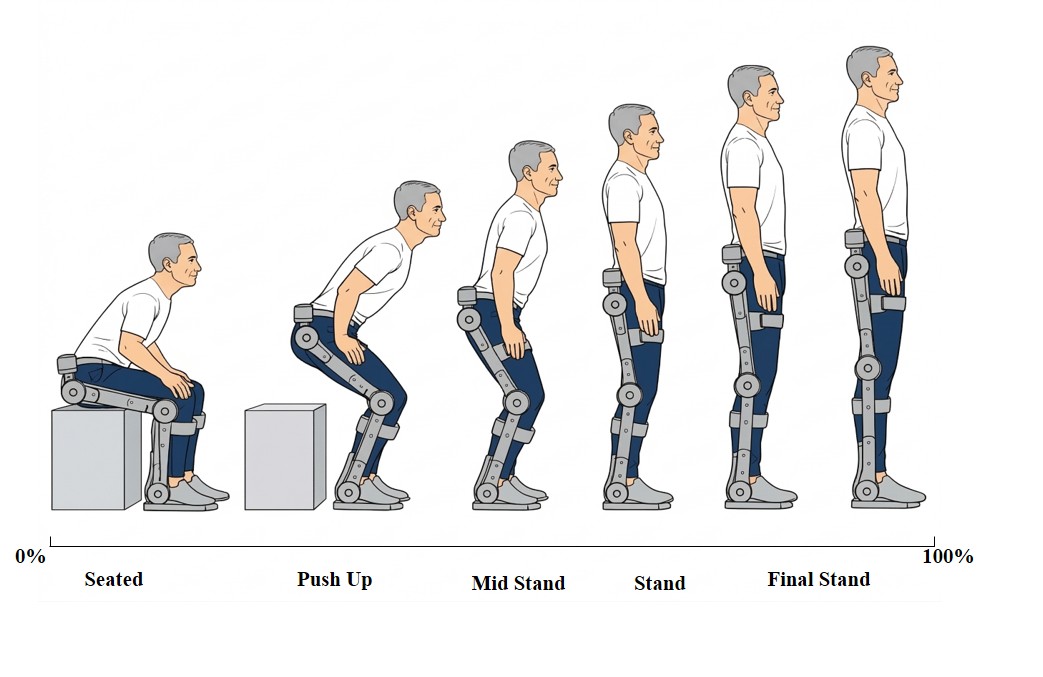}
\caption{Sequence of STS motion in the human exoskeleton system.}
\label{fig:fig2}
\end{figure}

The STS transition is a demanding daily activity that transfers the body weight from a supported seated posture to an upright standing posture. For elderly individuals, this motion is associated with significant muscle weakness and joint loading, making it a primary target for exoskeleton-assisted rehabilitation. As illustrated in Fig.~\ref{fig:fig2}, the STS motion can be decomposed into five distinct phases: Seated, Push-Up, Mid-Stand, Stand, and Final Stand, progressing from 0\% to 100\% of the motion cycle. Each phase is characterized by specific joint-angle configurations and varying levels of gravitational loading, with the push-up phase imposing the largest torque demands at the hip joint.

\begin{figure}[!ht]
\centering
\includegraphics[width=0.85\linewidth]{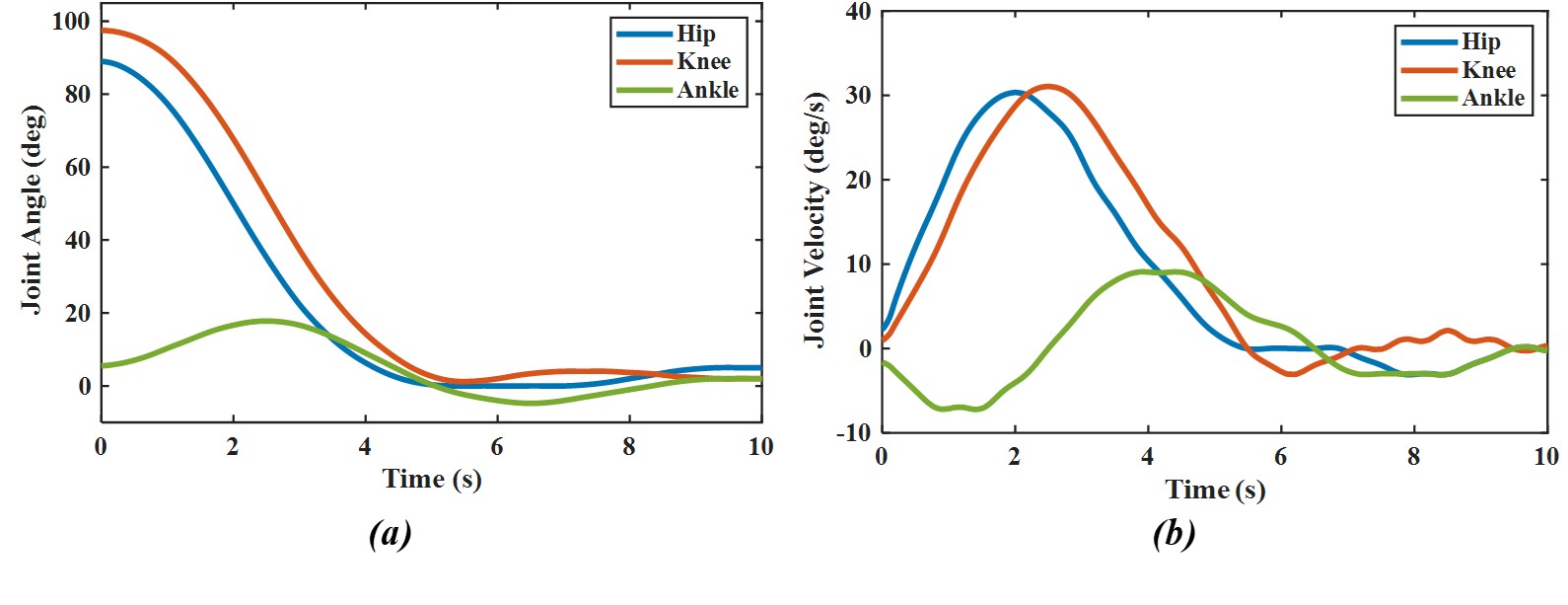}
\caption{OpenSim-derived reference joint kinematics for the STS motion: (a) joint angle trajectories; (b) joint angular velocity profiles.}
\label{fig:fig3}
\end{figure}

The reference joint trajectories used in this study were generated using OpenSim software inverse kinematics for a 50th-percentile reference subject with a body mass of 75~kg and height of 1.75~m, ensuring that the reference signals are representative of a typical healthy adult performing a standard STS motion. The raw kinematic data were low-pass filtered at 6~Hz to remove high-frequency noise artifacts, normalized to a duration of 10~s, and resampled at 1~kHz to produce compatible reference signals. The resulting dataset provides the desired joint angles and angular velocities for the hip, knee, and ankle joints, as shown in Fig.~\ref{fig:fig3}.

The reference trajectories depict the natural biomechanical progression of STS motion, with the hip and knee joints exhibiting the largest angular displacements from the fully seated to the fully upright configuration. The ankle joint provides a minor but important angular change to maintain balance throughout the rising motion. Since the STS task is performed symmetrically, identical reference trajectories are applied to the right and left limbs. These trajectories serve as the desired tracking signals for the proposed controller and all benchmark controllers evaluated in Section~\ref{sec:results}, ensuring consistent and fair performance comparisons with state-of-the-art controllers.

\subsection{Co-Simulation Framework}
\label{subsec:cosim}

\begin{figure}[!ht]
\centering
\includegraphics[width=0.75\linewidth]{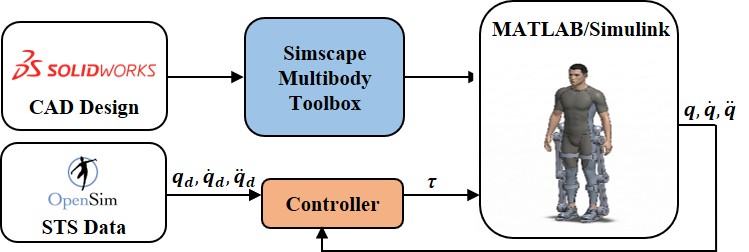}
\caption{Co-simulation framework for human exoskeleton system.}
\label{fig:fig4}
\end{figure}

This study includes a co-simulation framework that integrates SolidWorks, Simscape Multibody Toolbox, and OpenSim into a unified simulation environment for controller design and evaluation, as illustrated in Fig.~\ref{fig:fig4}. The exoskeleton mechanical assembly designed in SolidWorks was imported into Simscape Multibody, which automatically generates a dynamic simulation model that preserves all rigid-body transforms, joint definitions, and mass properties from the original CAD geometry. The reference joint trajectories generated from OpenSim inverse kinematics are fed as desired tracking signals to the controller, which computes actuator torques at each joint and applies them to the Simscape plant, closing the control loop.

The simulation employs the ode45 variable-step solver with a relative tolerance of $1 \times 10^{-3}$ to handle the closed kinematic loop constraints in the bilateral exoskeleton. All benchmark and proposed controllers are evaluated under identical plant configurations, solver settings, and reference trajectories to ensure a consistent and fair performance comparison.

\section{Proposed Control Design}
\label{sec:control}

\begin{figure}[!ht]
\centering
\includegraphics[width=0.95\linewidth]{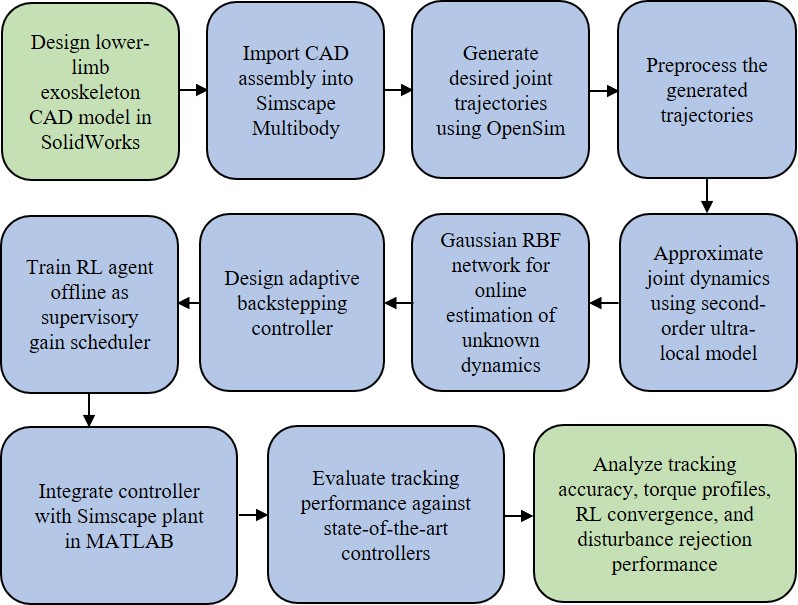}
\caption{Overall workflow of the proposed lower-limb exoskeleton control framework.}
\label{fig:fig5}
\end{figure}

This section presents the design of a model-free adaptive control strategy built upon a second-order model and a backstepping control framework. The overall methodology adopted in this study is illustrated in Fig.~\ref{fig:fig5}.

\begin{figure}[!ht]
\centering
\includegraphics[width=0.9\linewidth]{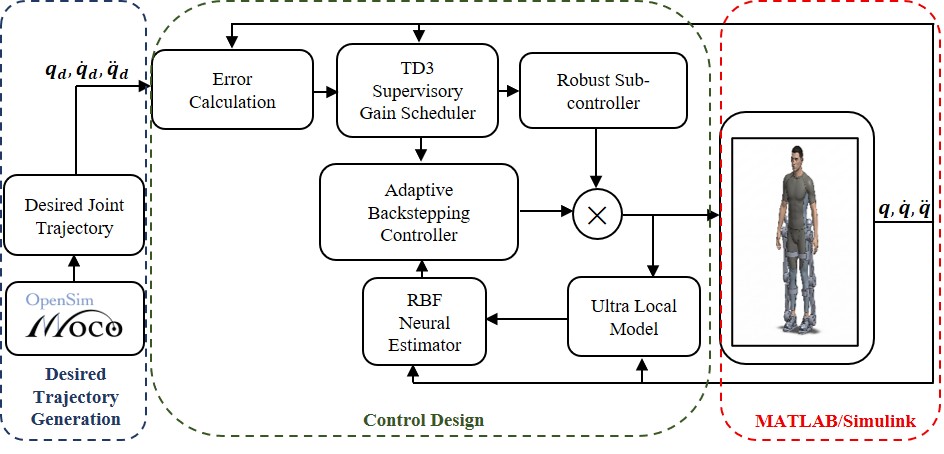}
\caption{Proposed control design architecture.}
\label{fig:fig6}
\end{figure}

Instead of relying on a full rigid-body dynamic model, which is inherently difficult to derive accurately due to the complexity and inter-subject variability of human--exoskeleton interaction, the ultra-local representation captures the essential input--output behavior of each joint without requiring prior knowledge of its physical parameters. The unknown lumped dynamics are estimated online using a Gaussian RBF neural network. To further enhance tracking performance across the distinct phases of STS motion, the TD3 agent is incorporated as a supervisory gain scheduler that adaptively adjusts selected controller gains in real time. The proposed control design architecture, shown in Fig.~\ref{fig:fig6}, integrates these components into a unified framework to ensure that the joint angles accurately track the desired STS trajectories.

\subsection{Problem Statement and Ultra-Local Model}
\label{subsec:ultra_local}

The coupled human--exoskeleton system performing STS motion can be mathematically formalized as an interconnected multi-body rigid system. By applying the Euler--Lagrange equations of motion, the overall system dynamics for three joints $j$ ($j=1$: hip, $j=2$: knee, $j=3$: ankle) are expressed as:
\begin{equation}
M(q_j)\,\ddot{q}_j + C(q_j,\dot{q}_j)\,\dot{q}_j + G(q_j) + \tau_d = \tau_j
\label{eq:dynamics}
\end{equation}

\noindent where $q_j$, $\dot{q}_j$, and $\ddot{q}_j$ represent the joint position, velocity, and acceleration, respectively. The term $M(q_j)$ denotes the inertia matrix, $C(q_j,\dot{q}_j)\,\dot{q}_j$ represents the Coriolis and centripetal forces, $G(q_j)$ is the gravitational torque vector, $\tau_d$ represents unknown disturbances, and $\tau_j$ is the applied actuator torque. In practice, controlling based on this dynamic model is complicated due to uncertainties, inter-subject variability, joint friction, and the difficulty of accurately identifying patient-specific parameters in a rehabilitation setting.

To overcome these limitations, the ultra-local model, introduced in the framework of model-free control, is adopted. This approach bypasses the need for an accurate mathematical model of the system by locally approximating each joint's input--output behavior. Instead of constructing a complete dynamic model based on physical laws and parameters, the system behavior within a sufficiently small duration window is captured by a simplified equation involving only measurable variables and an unknown term that lumps all unmodeled effects. For the lower-limb exoskeleton, a second-order ultra-local model is adopted to approximate the joint dynamics \citep{Fliess2013}:
\begin{equation}
\ddot{q} = F(t) + a\,\tau
\label{eq:ultra_local}
\end{equation}

\noindent where the parameter $a$ is a user-defined scaling coefficient chosen so that the magnitudes of $a\tau$ and $\ddot{q}$ are comparable. The term $F(t)$ is a continuously varying, unknown function that lumps together all unmodeled dynamics, including inertia, gravity, Coriolis effects, human interaction, and external disturbance.

The success of this model-free approach depends critically on the ability to estimate $F(t)$ accurately in real time. Once this estimation is available, it becomes possible to design effective feedback control laws without relying on system identification or dynamic modeling. However, since $F(t)$ aggregates all the unknown and time-varying dynamics of the human--exoskeleton system, estimating it poses a significant challenge. A neural network with strong nonlinear approximation capability is therefore required to capture this term reliably during the STS motion. To this end, a Gaussian radial basis function neural network is designed and integrated into the control framework.

\subsection{RBF Neural Estimator and Weight Adaptation Law}
\label{subsec:rbf}

\begin{figure}[!ht]
\centering
\includegraphics[width=0.6\linewidth]{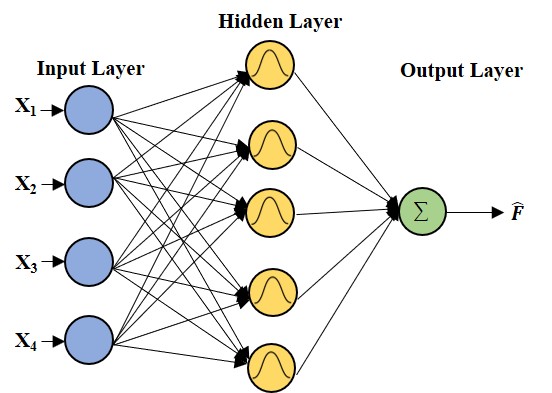}
\caption{Architecture of the Gaussian RBF neural network used for online lumped dynamics estimation.}
\label{fig:fig7}
\end{figure}

A successful model-free control strategy relies on accurate real-time estimation of the unknown lumped term $F(t)$. As it encapsulates highly nonlinear and time-varying dynamics of the human--exoskeleton system, a neural network with strong nonlinear approximation capability is required. In this study, a Gaussian RBF neural network is adopted to solve these challenges, as shown in Fig.~\ref{fig:fig7}. The RBF network is implemented independently for each joint $j$, and the following RBF neural estimator formulation is presented for a representative joint. It employs fixed neuron centers and adapts only the output weight vector, resulting in a simpler adaptation law, reduced computational load, and a cleaner stability proof \citep{Narayan2023}. The RBF network estimates $F(t)$ as:
\begin{equation}
\hat{F}(t) = \hat{W}^{\top} h(x)
\label{eq:rbf_estimate}
\end{equation}

\noindent where $\hat{W} \in \mathbb{R}^{n}$ is the adaptive output-weight vector, $h(x)$ is the RBF activation vector, and $n$ is the number of hidden nodes. The input to the RBF network is constructed from the measurable joint states $(q, \dot{q})$ and the desired reference $(q_d, \dot{q}_d)$ signals:
\begin{equation}
x = [\,q,\, \dot{q},\, q_d,\, \dot{q}_d\,]^{\top}
\label{eq:rbf_input}
\end{equation}

This input vector provides the network with sufficient information about both the current system state and the tracking reference, enabling accurate approximation of $F$ across all phases of the STS motion. For the $r$th hidden node, the Gaussian activation function is presented as \citep{Narayan2023}:
\begin{equation}
h_r(x) = \exp\!\left(-\frac{\|x - C_r\|^2}{2\sigma^2}\right)
\label{eq:rbf_activation}
\end{equation}

\noindent where $C_r$ denotes the fixed center vector of the $r$th hidden neuron, and $\sigma$ is the Gaussian width parameter used for all hidden neurons. The full basis activation vector is:
\begin{equation}
h(x) = [\,h_1(x),\, h_2(x),\, \ldots,\, h_n(x)\,]^{\top}
\label{eq:rbf_vec}
\end{equation}

By the universal approximation theorem \citep{Park1991} for RBF networks, for any continuous function $F$ on a compact set $\Omega_x$, there exist ideal weights $W^{*}$ and a bounded approximation residual $\varepsilon$ such that:
\begin{equation}
F(t) = W^{*\top} h(x) + \varepsilon, \qquad |\varepsilon| \le \varepsilon_N
\label{eq:approx_theorem}
\end{equation}

The estimation error between the true and estimated lumped dynamics can be represented by:
\begin{equation}
F(t) - \hat{F}(t) = \tilde{W}^{\top} h(x) + \varepsilon
\label{eq:estimation_error}
\end{equation}

\noindent where $\tilde{W} = W^{*} - \hat{W}$ is the weight estimation error. Since the RBF centers are fixed, Equation~(\ref{eq:estimation_error}) contains no error term arising from a drifting input-layer matrix. This is a key simplification over MLP-based estimators and directly enables the single-weight stability analysis. The adaptive output-weight vector $\hat{W}$ is updated online using a gradient-descent law with $\sigma$-modification leakage \citep{Moran2026}:
\begin{equation}
\dot{\hat{W}} = \lambda_A\, e_2\, h(x) - \eta\, \hat{W}
\label{eq:weight_update}
\end{equation}

\noindent where $\lambda_A$ is the adaptation gain, $e_2$ is the backstepping velocity error defined in Section~\ref{subsec:backstepping}, and $\eta$ is the leakage coefficient. In Equation~(\ref{eq:weight_update}), the term $\eta \hat{W}$ prevents weight drift during intervals of persistent estimation error or near-zero excitation, ensuring boundedness of $\hat{W}$. Initial weights are set to small random values. The RBF network parameters are summarized in Table~\ref{tab:rbf_params}.

\begin{table}[!ht]
\centering
\caption{Parameters of the RBF neural network.}
\label{tab:rbf_params}
\small
\begin{tabular}{@{}ll@{}}
\toprule
\textbf{Parameter} & \textbf{Value} \\
\midrule
RBF network type           & Fixed-center Gaussian RBF network \\
Input dimension            & 4 \\
Output dimension           & 1 \\
Network output             & Estimated lumped nonlinear dynamics \\
Number of hidden neurons   & 5 \\
Activation function        & Gaussian radial basis function \\
Gain scale range           & 0.50 to 1.50 \\
Gaussian width             & 2.236 \\
Adaptation learning rate   & 50 \\
\bottomrule
\end{tabular}
\end{table}

\subsection{Adaptive Backstepping Control Law}
\label{subsec:backstepping}

Based on the ultra-local model Equation~(\ref{eq:ultra_local}) and the RBF estimate $\hat{F}$ derived above, the control torque is synthesized using the backstepping method. Backstepping is a recursive Lyapunov-based design method particularly suited for systems in strict-feedback form, as it constructs a stabilizing control law step by step while guaranteeing closed-loop stability at each stage. Defining the system states $x_1 = q$ and $x_2 = \dot{q}$, the ultra-local model \citep{Fliess2013,Amara2025} is equivalently expressed as:
\begin{align}
\dot{x}_1 &= x_2 \label{eq:state1} \\
\dot{x}_2 &= F(t) + a\,\tau \label{eq:state2}
\end{align}

The backstepping design proceeds in two recursive steps. The position-tracking error between the actual and desired joint angle is defined as:
\begin{equation}
e_1 = x_1 - q_d
\label{eq:e1}
\end{equation}

The first Lyapunov candidate function is selected as:
\begin{equation}
V_1 = \frac{1}{2}\,e_1^{2}
\label{eq:V1}
\end{equation}

Taking the derivative of Equation~(\ref{eq:V1}) gives:
\begin{equation}
\dot{V}_1 = e_1\,\dot{e}_1 = e_1\,(x_2 - \dot{q}_d)
\label{eq:V1_dot}
\end{equation}

To make Equation~(\ref{eq:V1_dot}) negative definite, a virtual control law $\beta$ is introduced:
\begin{equation}
\beta = \dot{q}_d - k_1\,e_1
\label{eq:beta}
\end{equation}

\noindent where $k_1$ is a positive design gain. The second error variable is defined as:
\begin{equation}
e_2 = x_2 - \beta
\label{eq:e2}
\end{equation}

Substituting Equation~(\ref{eq:e2}) into Equation~(\ref{eq:V1_dot}) gives:
\begin{equation}
\dot{V}_1 = -k_1\,e_1^{2} + e_1\,e_2
\label{eq:V1_dot_sub}
\end{equation}

For the velocity error tracking, the augmented Lyapunov candidate function is chosen as:
\begin{equation}
V_2 = V_1 + \frac{1}{2}\,e_2^{2} + \frac{1}{2\lambda_A}\,\tilde{W}^{\top}\tilde{W}
\label{eq:V2}
\end{equation}

The derivative of Equation~(\ref{eq:V2}) gives:
\begin{equation}
\dot{V}_2 = \dot{V}_1 + e_2\,\dot{e}_2 + \frac{1}{\lambda_A}\,\tilde{W}^{\top}\dot{\tilde{W}}
\label{eq:V2_dot}
\end{equation}

The derivative of the virtual control is obtained analytically:
\begin{align}
\dot{\beta} &= \ddot{q}_d - k_1\,\dot{e}_1 \label{eq:beta_dot1} \\
            &= \ddot{q}_d - k_1\,(x_2 - \dot{q}_d) \label{eq:beta_dot2}
\end{align}

\noindent where $\ddot{q}_d$ is the desired acceleration. Replacing $F$ with $\hat{F}$, the backstepping control law is:
\begin{equation}
\tau_{\text{eq}} = \frac{1}{a}\Big[\ddot{q}_d - \hat{F} + k_1\,\dot{e}_1 - e_1 - k_2\,e_2\Big]
\label{eq:tau_eq}
\end{equation}

However, due to the residual neural approximation error and bounded external disturbances, the control law Equation~(\ref{eq:tau_eq}) alone cannot fully ensure robustness. To reduce these effects, a robust auxiliary sub-controller is included:
\begin{equation}
\tau_r = -\frac{1}{a}\,K_r\,\|h(x)\|^2\,e_2
\label{eq:tau_r}
\end{equation}

\noindent where $K_r$ is the robustness gain, and $\|h(x)\|^2$ denotes the squared Euclidean norm of the RBF activation vector. This term provides a state-dependent damping gain that enhances the robustness when the RBF network is strongly activated near a basis center. The total torque of the actuator before saturation is given by:
\begin{equation}
\tau_{\text{raw}} = \tau_{\text{eq}} + \tau_r
\label{eq:tau_raw}
\end{equation}

\noindent where $\tau_{\text{raw}}$ is the unsaturated torque command; the final saturated torque applied to the joint is:
\begin{equation}
\tau = \text{sat}(\tau_{\text{raw}}) = \max\!\big(\tau_{\min},\, \min(\tau_{\max},\, \tau_{\text{raw}})\big)
\label{eq:tau_sat}
\end{equation}

\noindent where $\tau_{\min}$ and $\tau_{\max}$ are the joint-specific lower and upper torque limits.

\subsection{TD3 Supervisory Gain Scheduler}
\label{subsec:td3}

To improve the adaptability of the RBF-MFAC backstepping controller, a reinforcement learning framework based on the TD3 algorithm is integrated as a supervisory-gain scheduler. The RBF-MFAC controller provides a stable low-level torque law, but its gains $k_1$, $k_2$, and $K_r$ are fixed parameters that may not be equally optimal during the STS phases, each of which imposes different dynamic loading and tracking demands on the joints. By replacing fixed-gain tuning with a learned policy, the TD3 agent continuously adapts controller gains based on observed joint states, desired trajectories, and tracking errors, thereby improving tracking accuracy while preserving the controller's stability structure.

\subsubsection{Reinforcement Learning Setup}

The RL environment is fundamental to the TD3 training process. The environment is intended to simulate the interaction between the lower limb exoskeleton and its control system during a 10~s STS episode. It includes the Simscape Multibody exoskeleton plant, the STS reference trajectories, and the RBF-MFAC backstepping controller as integral components of the environment. The TD3 agent interacts with the environment by receiving an observation vector at each agent step and generating a continuous gain scaling action that updates the controller gains to reduce tracking errors throughout the STS motion. These scaled gains are immediately passed into the controller, which calculates the final saturated actuator torque. An episode ends when the STS duration is complete or an unsafe tracking condition is detected.

\subsubsection{State Space}

The TD3 observation is defined as a 6-dimensional continuous state vector constructed independently for each joint agent at each execution step:
\begin{equation}
s_t = [\,e_1,\, e_2,\, q,\, q_d,\, \tau_{t-1},\, \rho\,]^{\top}
\label{eq:state_vec}
\end{equation}

The state space includes position and velocity tracking errors, actual and desired joint positions, the previous actuator torque $\tau_{t-1}$, and the normalized STS movement phase $\rho$, given as:
\begin{equation}
\rho = \frac{t}{T}, \qquad \rho \in [0, 1]
\label{eq:phase}
\end{equation}

\noindent where $T$ is the episode duration. The state vector provides the agent with direct tracking error information, joint state information, actuator history, and task phase information, enabling phase-aware gain adaptation throughout the STS motion. As three independent TD3 agents are used for the three joints, the combined framework includes 18 state variables.

\subsubsection{Action Space}

The TD3 action is a 3-dimensional continuous vector that can be represented as:
\begin{equation}
a_t = [\,u_{k_1},\, u_{k_2},\, u_{K_r}\,]^{\top} \in [-1, 1]^{3}
\label{eq:action}
\end{equation}

Each action component represents a normalized gain-scaling command. A linear mapping converts $u_i$ to a scale factor:
\begin{equation}
\text{scale}_i = 1 + 0.5\,u_i, \qquad \text{scale}_i \in [0.5, 1.5]
\label{eq:scale}
\end{equation}

Therefore, the effective controller gains applied to the RBF-MFAC backstepping controller are presented as:
\begin{equation}
k_1 = \text{scale}_1\,k_{01}, \quad k_2 = \text{scale}_2\,k_{02}, \quad K_r = \text{scale}_3\,k_{03}
\label{eq:gains}
\end{equation}

\noindent where $k_{01}$, $k_{02}$, and $k_{03}$ are the nominal baseline gains. The gain scale range keeps all gains positive and within a safe operating band, preserving stability. The remaining controller parameters remain unchanged.

\subsubsection{Reward Function}

In RL, the reward function defines the goal that the agent attempts to achieve through its actions. For the STS gain-scheduling task, the reward function is designed to balance two competing goals: minimizing tracking error and limiting control effort. The reward function is defined as:
\begin{equation}
r_t = -\big(w_p\,e_1^{2} + w_v\,e_2^{2} + w_\tau\,\tau^{2}\big)
\label{eq:reward}
\end{equation}

\noindent where $w_p$ penalizes position tracking error, $w_v$ penalizes backstepping velocity error, and $w_\tau$ penalizes large actuator torque. Since all terms are negative, reward values near zero indicate good tracking performance with minimal actuator effort.

\subsubsection{TD3 Algorithm}

\begin{figure}[!ht]
\centering
\includegraphics[width=0.7\linewidth]{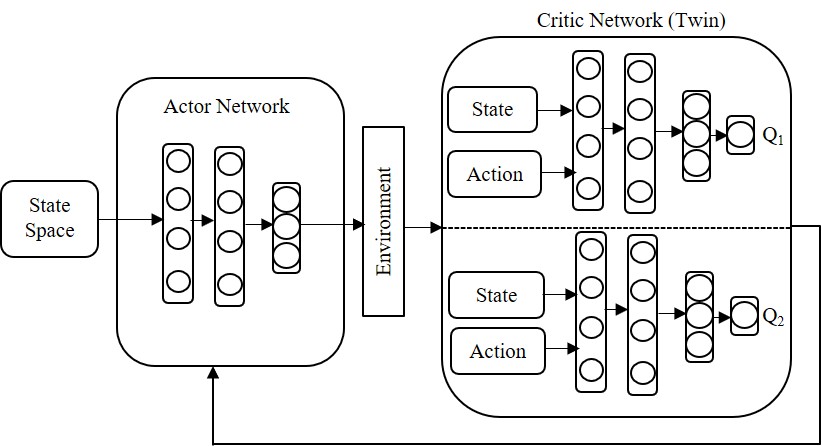}
\caption{Twin Delayed Deep Deterministic Policy Gradient (TD3) architecture.}
\label{fig:fig8}
\end{figure}

TD3 is an off-policy actor--critic algorithm for continuous control that extends DDPG by addressing Q-value overestimation and training instability through three mechanisms: clipped double-Q learning, delayed policy updates, and target policy smoothing \citep{Fujimoto2018}. The actor network $\mu_\theta(s_t)$ transforms the state into a deterministic continuous action, while two independent critic networks $Q_{\varphi_1}(s_t, a_t)$ and $Q_{\varphi_2}(s_t, a_t)$ estimate the action-value function. Transitions $(s_t, a_t, r_t, s_{t+1})$ are stored in a replay buffer and sampled uniformly for network updates. Figure~\ref{fig:fig8} illustrates the overall TD3 architecture adopted for the supervisory gain-scheduling framework.

To prevent overestimation of target values, the minimum of the two target critic estimates is used when computing the TD3 target \citep{Fujimoto2018}:
\begin{equation}
y_t = r_t + \gamma\,(1 - d_t)\,\min_{i=1,2} Q_{\varphi_i'}\!\big(s_{t+1},\, \tilde{a}\big)
\label{eq:td3_target}
\end{equation}

\noindent where $\gamma$ is the discount factor and $d_t$ is the terminal-state indicator. Each critic is trained by minimizing:
\begin{equation}
\mathcal{L}(\varphi_i) = \mathbb{E}\!\left[\big(Q_{\varphi_i}(s_t, a_t) - y_t\big)^{2}\right]
\label{eq:critic_loss}
\end{equation}

In the delayed policy update, the actor is updated less frequently than the critics, ensuring that value estimates are sufficiently accurate before the policy is updated \citep{Fujimoto2018}:
\begin{equation}
\nabla_\theta J(\theta) = \mathbb{E}\!\left[\nabla_a Q_{\varphi_1}(s, a)\big|_{a=\mu_\theta(s)} \nabla_\theta \mu_\theta(s)\right]
\label{eq:actor_grad}
\end{equation}

Both the actor and critic target networks are updated via soft updates:
\begin{equation}
\varphi_i' \leftarrow \xi\,\varphi_i + (1-\xi)\,\varphi_i', \qquad \theta' \leftarrow \xi\,\theta + (1-\xi)\,\theta'
\label{eq:soft_update}
\end{equation}

\noindent where $\xi$ is the target-smoothing factor. In target policy smoothing \citep{Fujimoto2018}, clipped Gaussian noise is added to the target action to regularize the critic and prevent overfitting to sharp action-value peaks:
\begin{equation}
\tilde{a} = \mu_{\theta'}(s_{t+1}) + \text{clip}\!\big(\mathcal{N}(0, \tilde{\sigma}^{2}),\, -c,\, c\big)
\label{eq:target_smooth}
\end{equation}

The small clipped noise added to the target action forces the critic to generate consistent value estimates across a neighborhood of actions rather than fitting to narrow action-value peaks. This regularization improves generalization of the critic and produces a smoother, more robust learned policy.

\subsection{Implementation Details}
\label{subsec:implementation}

The proposed controller was implemented and evaluated within a co-simulation environment using MATLAB R2025a and Simulink Simscape Multibody, as described in Section~\ref{subsec:cosim}. In Simulink, the controllers were applied to each joint separately with corresponding input and output signals. The controller parameters used in the proposed controller are listed in Table~\ref{tab:proposed_params}.

\begin{table}[!ht]
\centering
\caption{Control parameters of the proposed control design.}
\label{tab:proposed_params}
\small
\begin{tabular}{@{}lcccccc@{}}
\toprule
\textbf{Joint} & $a$ & $k_1$ & $k_2$ & $K_r$ & $\lambda$ & $\eta$ \\
\midrule
Hip   & 100 & 230 & 175 & 20 & 130 & 0.22 \\
Knee  & 50  & 150 & 100 & 15 & 50  & 0.10 \\
Ankle & 25  & 50  & 50  & 5  & 20  & 0.10 \\
\bottomrule
\end{tabular}
\end{table}

For the TD3 based supervisory gain scheduler, three independent agents were developed for the actuated joints (hip, knee, and ankle) using the MATLAB Reinforcement Learning Toolbox. Three separate TD3 agents, one per joint, were trained offline in Simulink before integrating with the proposed controller. The complete TD3 agent and training hyperparameters are listed in Table~\ref{tab:td3_params}.

\begin{table}[!ht]
\centering
\caption{TD3 agent training and network parameters.}
\label{tab:td3_params}
\small
\begin{tabular}{@{}ll@{}}
\toprule
\textbf{Parameter} & \textbf{Value} \\
\midrule
Agent type                       & TD3 \\
Role                             & Gain Scheduler \\
Action dimension                 & 3 \\
Action vector                    & $[u_{k_1}, u_{k_2}, u_{K_r}]$ \\
Action range                     & $[-1, 1]$ \\
Scaled gains                     & $k_1, k_2, K_r$ \\
Gain scale range                 & 0.50 to 1.50 \\
Observation dimension            & 6 \\
\midrule
\multicolumn{2}{l}{\textit{Training parameters}} \\
Maximum steps per episode        & 500 \\
Discount factor                  & 0.995 \\
Target smoothing factor          & 0.005 \\
Mini-batch size                  & 256 \\
Sample time                      & 0.02 \\
\midrule
\multicolumn{2}{l}{\textit{Exploration and policy noise}} \\
Exploration noise std.           & 0.30 \\
Exploration noise decay rate     & $1 \times 10^{-5}$ \\
Target policy smoothing noise std.& 0.20 \\
Target policy noise clipping range& $[-0.5, 0.5]$ \\
\midrule
\multicolumn{2}{l}{\textit{Actor--critic network configuration}} \\
Actor hidden layers              & $[128, 128]$ \\
Critic hidden layers             & $[128, 128]$ \\
Actor output activation          & tanh \\
Number of critic networks        & 2 \\
Actor learning rate              & $1 \times 10^{-3}$ \\
Critic learning rate             & $1 \times 10^{-3}$ \\
Optimizer                        & Adam \\
\bottomrule
\end{tabular}
\end{table}

\section{Results and Discussion}
\label{sec:results}

In this section, the performance of the proposed control algorithm is presented through simulation experiments alongside four benchmark strategies: PID, LQR, SMC, and MFAC, using a similar co-simulation configuration as described in Section~\ref{subsec:cosim}. All benchmark controllers were manually tuned under the same simulation environment to achieve their best possible tracking performance, and the corresponding controller parameters are summarized in Table~\ref{tab:benchmark_params}. The benchmark SMC employed a conventional sliding surface defined as:
\begin{equation}
s = \dot{e} + \lambda\,e
\label{eq:smc_surface}
\end{equation}

\noindent and the switching control law can be represented as:
\begin{equation}
u = u_{\text{eq}} - K\,\text{sat}\!\left(\frac{s}{\phi}\right)
\label{eq:smc_law}
\end{equation}

\noindent where $\lambda$, $K$, and $\phi$ denote the sliding-surface coefficient, switching gain, and boundary-layer thickness, respectively. The saturation function $\text{sat}(\cdot)$ replaces the discontinuous $\text{sign}(\cdot)$ function to mitigate chattering, producing a continuous control output within the boundary layer.

\begin{table}[!ht]
\centering
\caption{Controller tuning parameters for benchmark controllers.}
\label{tab:benchmark_params}
\small
\begin{tabular}{@{}lllll@{}}
\toprule
\textbf{Joint} & \textbf{PID ($K_p, K_i, K_d$)} & \textbf{LQR ($Q_\text{pos}, Q_\text{vel}, R$)} & \textbf{SMC ($\lambda, K, \Phi$)} & \textbf{MFAC ($a, k_1, k_2, K_r, \lambda, \eta$)} \\
\midrule
Hip   & $\{300, 60, 20\}$ & $\{4000, 50, 0.01\}$ & $\{3, 30, 0.17\}$  & $\{100, 163, 123, 15, 110, 0.30\}$ \\
Knee  & $\{150, 25, 15\}$ & $\{2000, 50, 0.01\}$ & $\{5, 20, 0.17\}$  & $\{50, 100, 80, 15, 50, 0.10\}$ \\
Ankle & $\{100, 20, 10\}$ & $\{1000, 15, 0.1\}$  & $\{8, 10, 0.15\}$  & $\{25, 50, 60, 5, 20, 0.10\}$ \\
\bottomrule
\end{tabular}
\end{table}

To enable a fair assessment of the contribution of the TD3 supervisory mechanism, the MFAC benchmark uses the same ultra-local backstepping architecture and online RBF neural-estimation framework as the proposed controller. The only difference is the exclusion of the TD3-based gain scheduler. Furthermore, the controllers are evaluated using the five metrics listed in Table~\ref{tab:metrics}. The reported results include joint tracking behavior, torque responses, TD3 training convergence, and finally, robustness of the proposed controller under external torque disturbances.

\begin{table}[!ht]
\centering
\caption{Performance metrics used to evaluate controllers.}
\label{tab:metrics}
\small
\begin{tabular}{@{}ll@{}}
\toprule
\textbf{Metric} & \textbf{Formula} \\
\midrule
Mean Absolute Error (MAE)        & $\displaystyle \text{MAE} = \frac{1}{N}\sum_{k=1}^{N} |e(k)|$ \\[6pt]
Root Mean Square Error (RMSE)    & $\displaystyle \text{RMSE} = \sqrt{\frac{1}{N}\sum_{k=1}^{N} e(k)^{2}}$ \\[6pt]
Peak Error                       & $\displaystyle e_{\text{peak}} = \max_{k} |e(k)|$ \\[6pt]
Root Mean Square Torque          & $\displaystyle \tau_{\text{RMS}} = \sqrt{\frac{1}{N}\sum_{k=1}^{N} \tau(k)^{2}}$ \\[6pt]
Peak Torque                      & $\displaystyle \tau_{\text{peak}} = \max_{k} |\tau(k)|$ \\
\bottomrule
\end{tabular}
\end{table}

\subsection{Joint Tracking Performance}
\label{subsec:tracking}

\begin{figure}[!ht]
\centering
\includegraphics[width=0.85\linewidth]{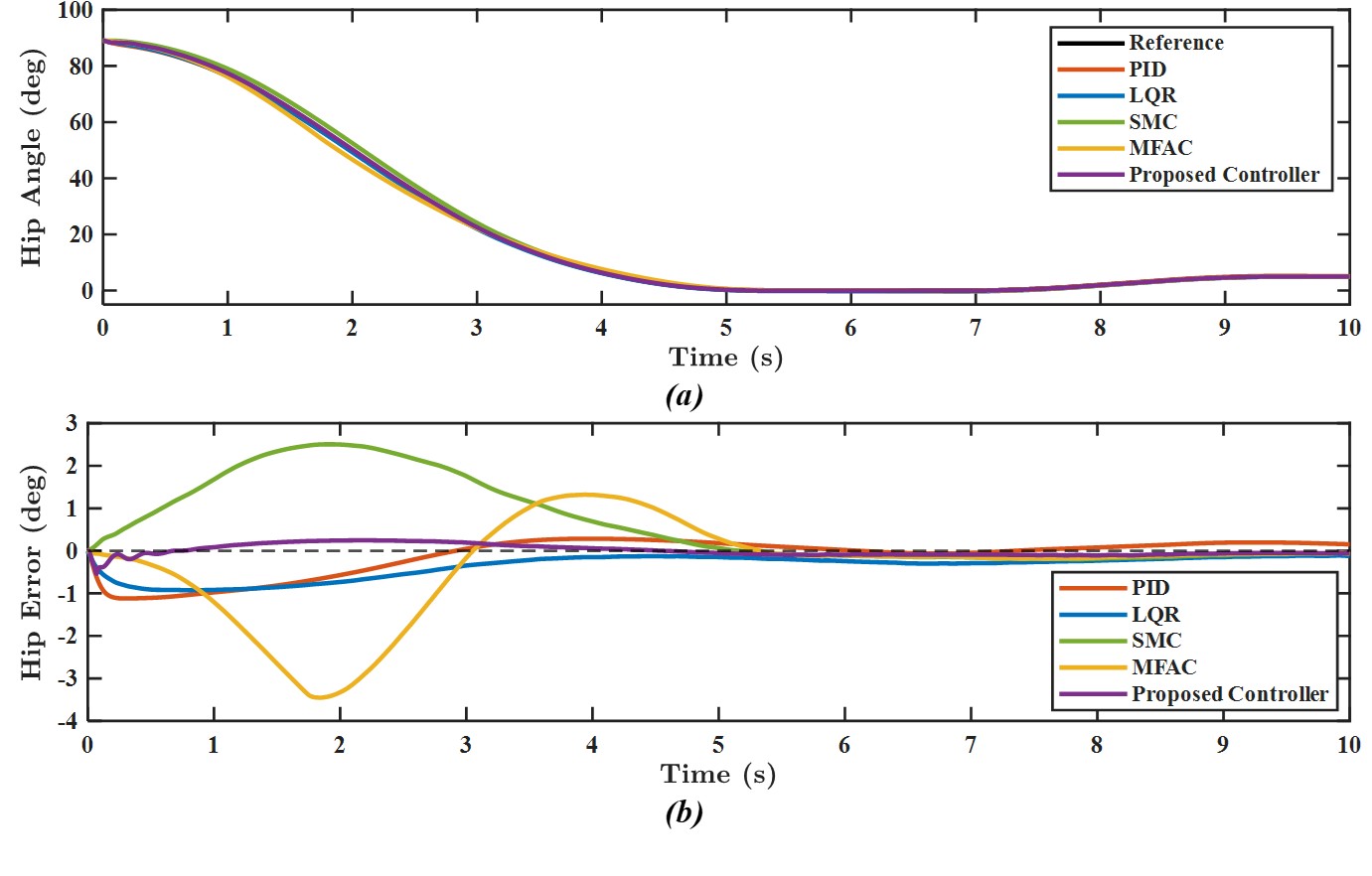}
\caption{Hip joint angle tracking performance during STS motion: (a) joint angle tracking; (b) tracking error.}
\label{fig:fig9}
\end{figure}

\begin{figure}[!ht]
\centering
\includegraphics[width=0.85\linewidth]{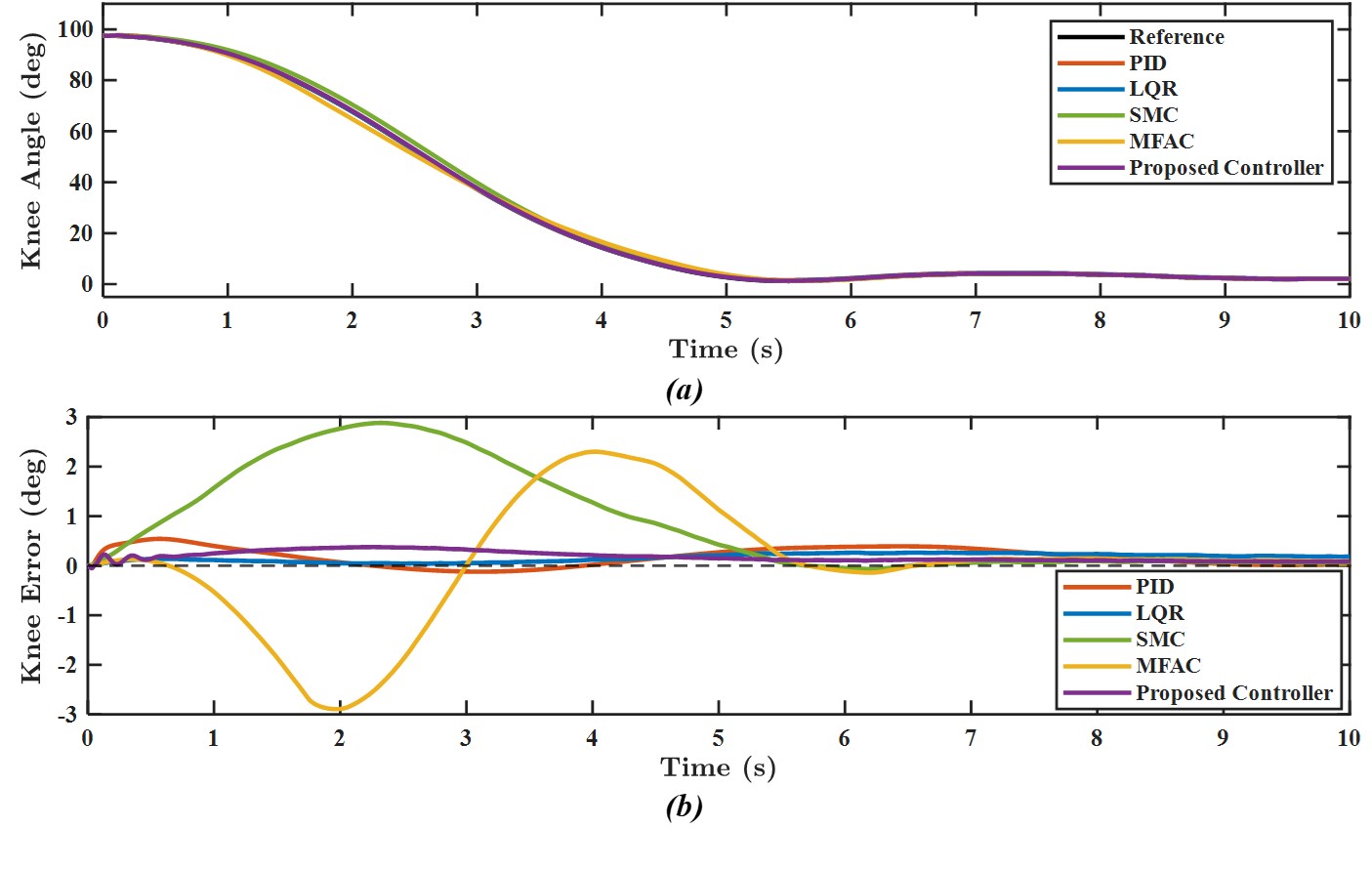}
\caption{Knee joint angle tracking performance during STS motion: (a) joint angle tracking; (b) tracking error.}
\label{fig:fig10}
\end{figure}

\begin{figure}[!ht]
\centering
\includegraphics[width=0.85\linewidth]{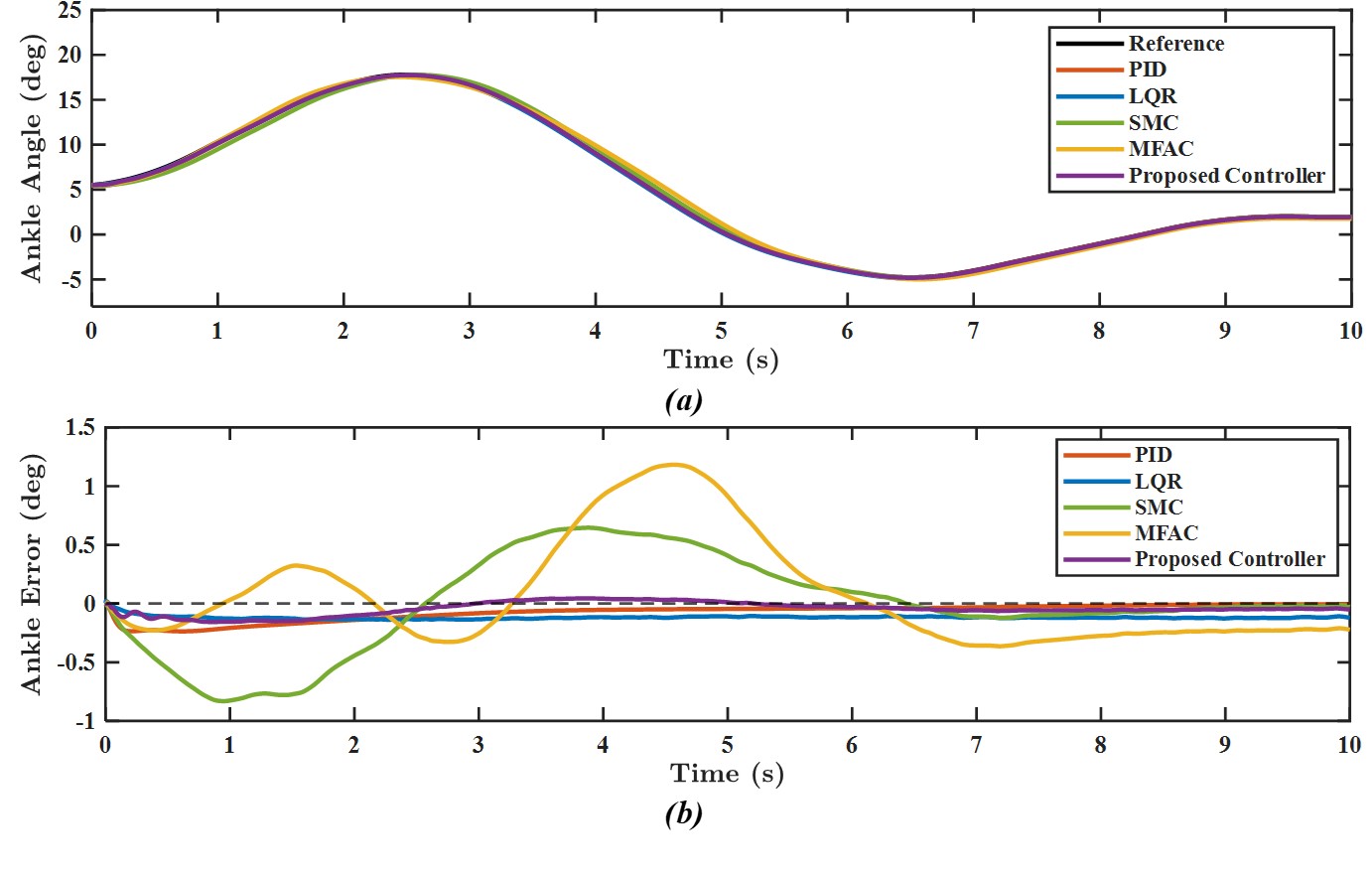}
\caption{Ankle joint angle tracking performance during STS motion: (a) joint angle tracking; (b) tracking error.}
\label{fig:fig11}
\end{figure}

The joint-angle tracking and its corresponding error profiles for all five controllers over the 10~s STS motion are presented in Figs.~\ref{fig:fig9}--\ref{fig:fig11}. As shown in Fig.~\ref{fig:fig9}, all controllers significantly follow the desired hip trajectory from the seated to the upright position. However, noticeable differences can be observed in the tracking-error responses. The MFAC controller has the highest negative peak error during the push-up phase, whereas the SMC shows moderate errors. PID and LQR maintain lower errors, while the proposed controller achieves near-zero tracking error throughout the motion.

Figure~\ref{fig:fig10} shows that the knee joint experiences the largest angular displacement during STS, making it the most demanding joint to control. Nearly all controllers closely followed the desired trajectory, but noticeable differences were observed in their tracking-error responses. The SMC controllers show the highest positive peak error, while the MFAC controller produces the largest negative tracking deviation during the transition phase. Although the PID and LQR controllers maintain comparatively smaller errors, noticeable deviations from the reference trajectory are still present. In contrast, the proposed controller consistently reports a tracking error close to zero, demonstrating superior tracking accuracy and smoother transient behavior of the knee joint.

Figure~\ref{fig:fig11} presents the ankle joint tracking results during the STS motion. Because of its smaller angular excursion and lower actuation requirements, the ankle joint is easier to control than the hip and knee joints. As observed from the tracking and error profiles, all controllers successfully follow the reference trajectory with small deviations. However, both MFAC and SMC show significant deviation and oscillatory behavior during the transition phase. While other controllers report reasonably low tracking errors, the proposed controller achieves the smallest error magnitude throughout, delivering improved precision and stability.

\subsection{Quantitative Comparison}
\label{subsec:quantitative}

The quantitative performance of all five controllers across the hip, knee, and ankle joints is summarized in Table~\ref{tab:perf_comparison}. The proposed controller reports the lowest average RMSE across all joints, representing reductions of 60.2\%, 48.7\%, 42.6\%, and 54.4\% compared to PID, LQR, SMC, and MFAC, respectively. Among the benchmark controllers, SMC provides the best overall tracking performance with an average RMSE of $0.136^{\circ}$, followed by MFAC ($0.171^{\circ}$), LQR ($0.152^{\circ}$), and PID ($0.196^{\circ}$).

\begin{table}[!ht]
\centering
\caption{Performance comparison of controllers.}
\label{tab:perf_comparison}
\small
\begin{tabular}{@{}llcccc@{}}
\toprule
\textbf{Controller} & \textbf{Joint} & \textbf{RMSE (deg)} & \textbf{Peak Error (deg)} & \textbf{RMS Torque (Nm)} & \textbf{Peak Torque (Nm)} \\
\midrule
\multirow{4}{*}{PID}
& Hip     & 0.354 & 0.768 & 92.189  & 145.021 \\
& Knee    & 0.123 & 0.192 & 15.802  & 191.54  \\
& Ankle   & 0.112 & 0.113 & 32.142  & 90.435  \\
& Average & 0.196 & 0.358 & --      & --      \\
\midrule
\multirow{4}{*}{LQR}
& Hip     & 0.255 & 0.390 & 110.154 & 151.043 \\
& Knee    & 0.094 & 0.086 & 34.864  & 59.398  \\
& Ankle   & 0.108 & 0.124 & 18.117  & 27.737  \\
& Average & 0.152 & 0.200 & --      & --      \\
\midrule
\multirow{4}{*}{SMC}
& Hip     & 0.244 & 0.551 & 26.206  & 30.955  \\
& Knee    & 0.081 & 0.079 & 11.355  & 20.424  \\
& Ankle   & 0.082 & 0.030 & 9.916   & 10.037  \\
& Average & 0.136 & 0.220 & --      & --      \\
\midrule
\multirow{4}{*}{MFAC}
& Hip     & 0.192 & 0.393 & 125.018 & 171.463 \\
& Knee    & 0.096 & 0.144 & 42.920  & 59.237  \\
& Ankle   & 0.224 & 0.256 & 14.678  & 20.253  \\
& Average & 0.171 & 0.264 & --      & --      \\
\midrule
\multirow{4}{*}{\textbf{Proposed}}
& Hip     & \textbf{0.124} & 0.557 & \textbf{17.591} & \textbf{28} \\
& Knee    & \textbf{0.064} & 0.197 & \textbf{5.984}  & \textbf{13} \\
& Ankle   & \textbf{0.047} & 0.079 & \textbf{2.981}  & \textbf{6}  \\
& Average & \textbf{0.078} & 0.278 & --      & --      \\
\bottomrule
\end{tabular}
\end{table}

\begin{figure}[!ht]
\centering
\includegraphics[width=0.85\linewidth]{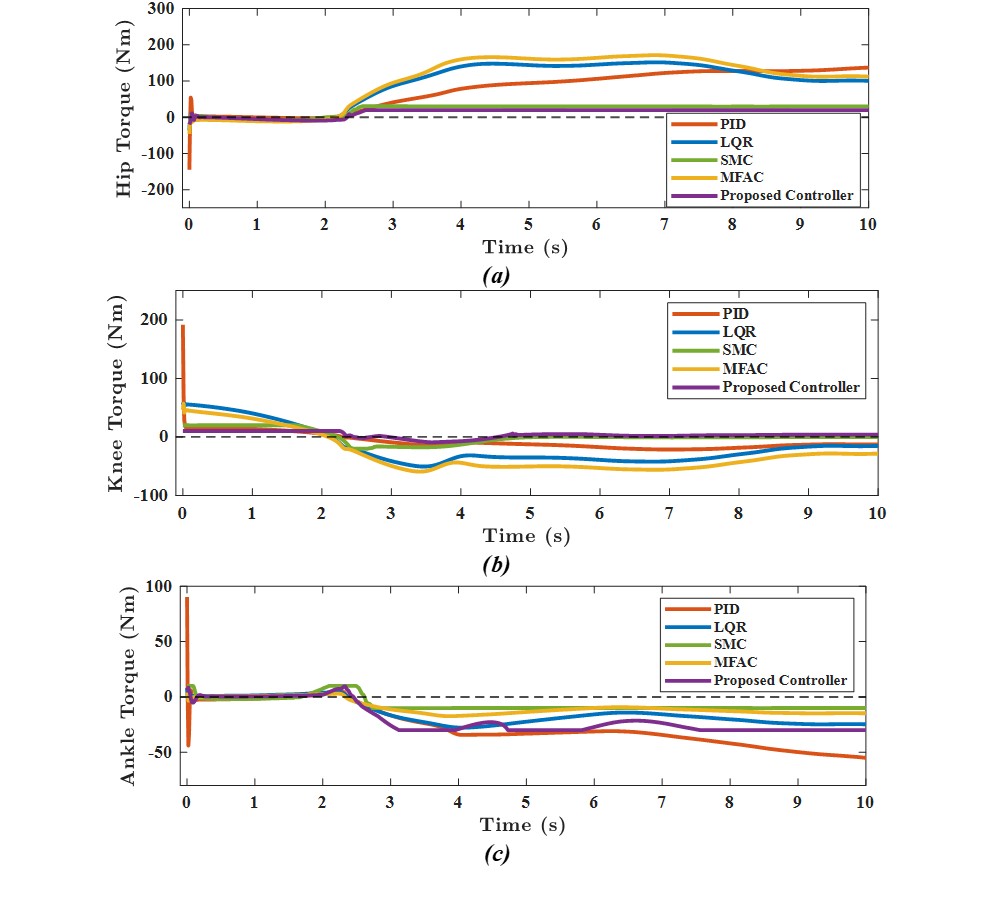}
\caption{Control torque profiles applied by each controller at the joints during STS motion: (a) hip; (b) knee; (c) ankle.}
\label{fig:fig12}
\end{figure}

In the torque analysis, the proposed controller also demonstrates the lowest actuator effort across all joints, with hip RMS torque of 17.591~Nm and peak torque of 28~Nm, significantly lower than PID and MFAC. This reduction in torque demand can be attributed to online RBF disturbance estimation, which directly compensates for gravity and inertia effects rather than relying on high-gain corrective measures. The torque results shown in Fig.~\ref{fig:fig12} further confirm that the proposed controller maintains lower torque commands throughout the STS motion at all three joints, whereas PID, LQR, and MFAC require significantly larger and more sustained torque efforts, particularly at the hip joint during the push-up phase.

\begin{figure}[!ht]
\centering
\includegraphics[width=0.65\linewidth]{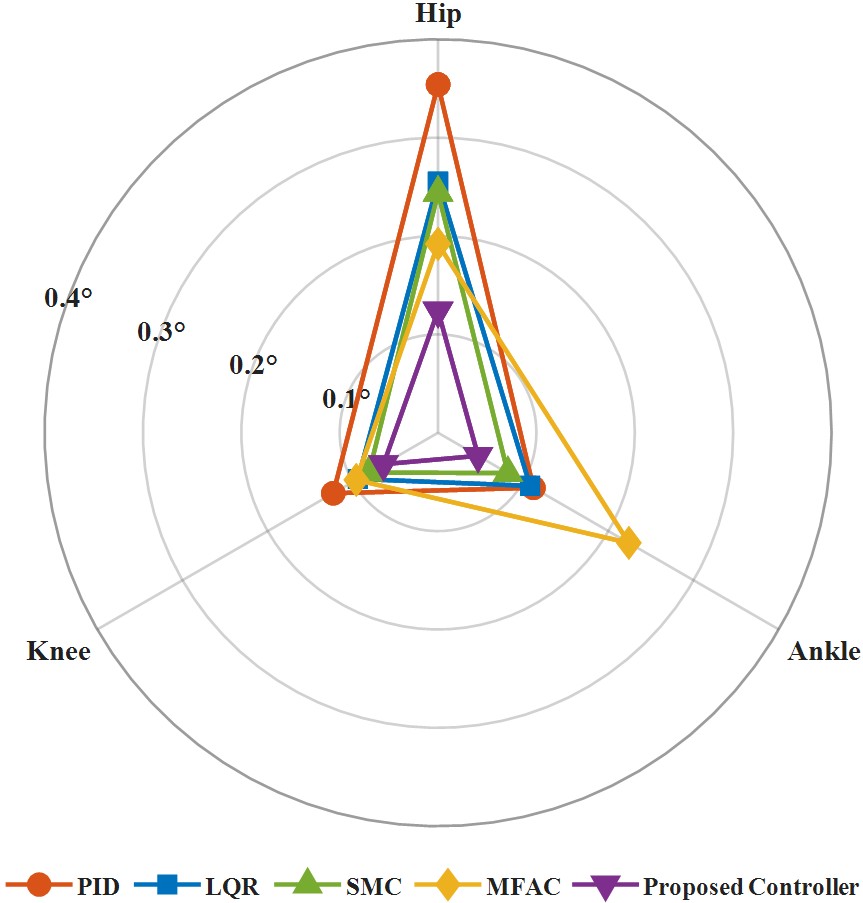}
\caption{Joint-wise RMSE comparison across hip, knee, and ankle for all five evaluated control strategies.}
\label{fig:fig13}
\end{figure}

The RMSE comparison presented in Fig.~\ref{fig:fig13} provides a concise visual assessment of tracking accuracy across all three joints simultaneously. The proposed controller consistently occupies the innermost region of the radar chart, indicating the lowest RMSE values of all evaluated controllers and confirming its superior tracking performance across all joints.

\subsection{TD3 Performance}
\label{subsec:td3_performance}

\begin{figure}[!ht]
\centering
\includegraphics[width=0.95\linewidth]{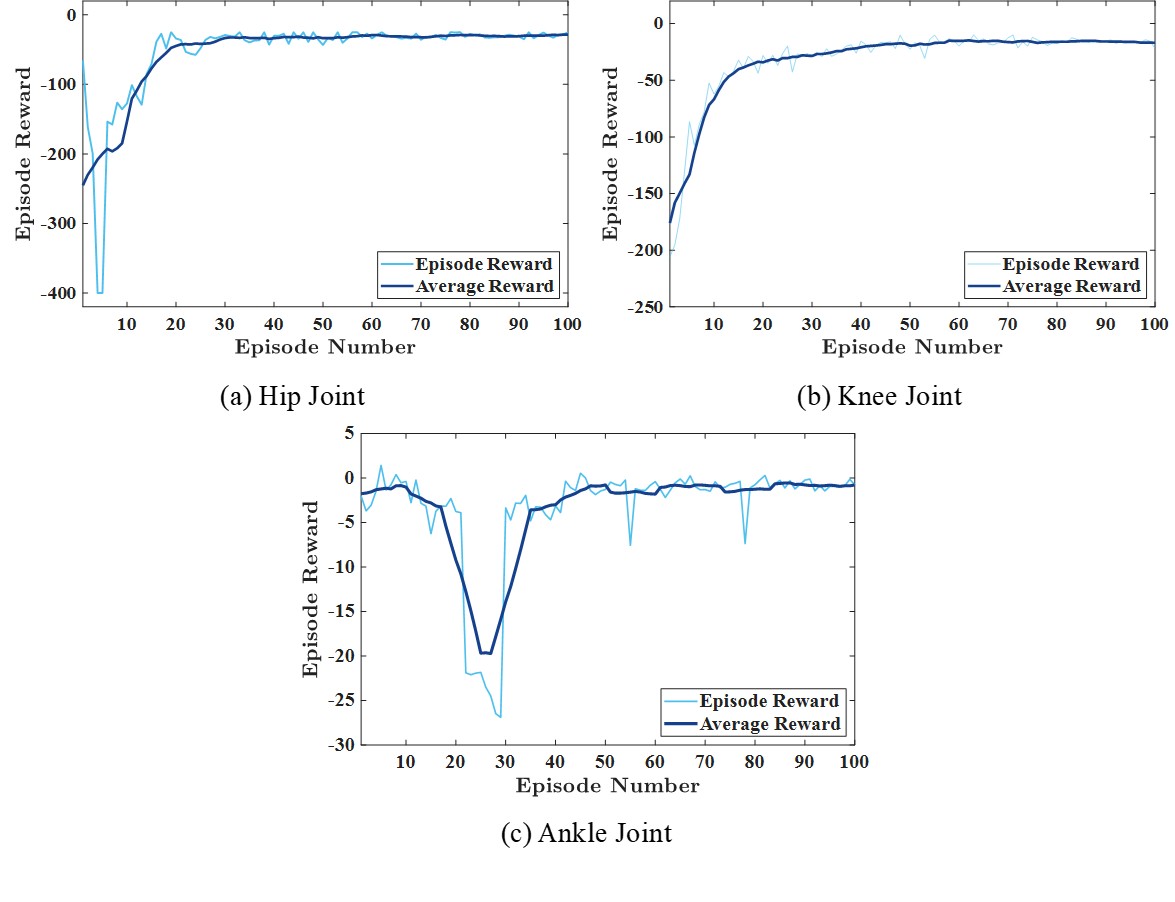}
\caption{TD3 agent training convergence curves for the proposed controller showing episode reward and average reward over 100 training episodes for (a) hip, (b) knee, and (c) ankle joints.}
\label{fig:fig14}
\end{figure}

Figure~\ref{fig:fig14} presents the TD3 agent training convergence curves for all three joints over 100 training episodes, including the episode reward and smoothed average reward at each joint. As shown in Fig.~\ref{fig:fig14}(a), the hip joint agent initially experiences large reward fluctuations during the exploration phase. The average reward increases rapidly from episode 1 to 20, indicating rapid initial learning due to the large hip tracking errors. The reward converges around episode 30--40 and remains consistent through episode 100, confirming the successful convergence of the hip gain-scheduling policy.

Figure~\ref{fig:fig14}(b) shows the knee joint training curve, which exhibits the smoothest convergence among the three joints and shows relatively small fluctuations throughout episodes. This smooth convergence reflects the relatively consistent dynamic loading at the knee joint during the STS push-up phase. Figure~\ref{fig:fig14}(c) presents the ankle joint training curve, which operates on a significantly smaller reward scale. A temporary reward dip is observed around episodes 25--30, after which the agent recovers and converges near zero by episode 50, indicating satisfactory learning performance with occasional exploration-induced fluctuations.

The performance improvement achieved through TD3 integration is summarized in Table~\ref{tab:td3_improvement} by comparing the standalone RBF-MFAC controller with the proposed controller design. The TD3 integration reports RMSE reductions of approximately 35\%, 33\%, and 79\% at the hip, knee, and ankle joints, respectively. Additionally, improvements in MAE are observed across the three joints, with the ankle benefiting most from adaptive gain scheduling. However, an increase in peak tracking error is observed at the hip and knee joints, suggesting a trade-off between reducing the average tracking error and limiting transient deviations. Since the TD3 reward function primarily focuses on optimizing cumulative tracking performance and control effort, peak-error behavior is not explicitly penalized. Overall, TD3 integration consistently improves average tracking accuracy across all joints.

\begin{table}[!ht]
\centering
\caption{Improvement in controller performance through TD3-assisted gain adaptation.}
\label{tab:td3_improvement}
\small
\begin{tabular}{@{}lccccccccc@{}}
\toprule
\textbf{Metric} & \multicolumn{3}{c}{\textbf{RMSE (deg)}} & \multicolumn{3}{c}{\textbf{MAE (deg)}} & \multicolumn{3}{c}{\textbf{Peak Error (deg)}} \\
\cmidrule(lr){2-4} \cmidrule(lr){5-7} \cmidrule(lr){8-10}
\textbf{Joint} & Hip & Knee & Ankle & Hip & Knee & Ankle & Hip & Knee & Ankle \\
\midrule
RBF-MFAC (without TD3)     & 0.192 & 0.096 & 0.224 & 0.158 & 0.082 & 0.185 & 0.393 & 0.144 & 0.256 \\
RBF-MFAC + TD3 (Proposed)  & 0.124 & 0.064 & 0.047 & 0.103 & 0.071 & 0.046 & 0.557 & 0.197 & 0.079 \\
Improvement (\%)           & 35.42 & 33.34 & 79.02 & 34.81 & 13.41 & 75.14 & $-$41.73 & $-$36.81 & 69.14 \\
\bottomrule
\end{tabular}
\end{table}

\subsection{Robustness Analysis}
\label{subsec:robustness}

To evaluate the robustness of the proposed controller against external disturbances, individual torque disturbances were applied separately to the hip, knee, and ankle joints during the STS motion. In each test, a constant disturbance torque was introduced at the midpoint of the simulation ($t = 5$~s) and maintained for a duration of 1~s. External disturbance magnitudes ranging from 0 to 12~Nm were applied, and the corresponding resulting RMSE values for each joint are summarized in Table~\ref{tab:robustness}. The results show that the proposed controller maintains satisfactory tracking performance despite increasing disturbance magnitudes. For the hip joint, the RMSE values increase gradually from $0.124^{\circ}$ under nominal conditions to $0.285^{\circ}$ under a 12~Nm disturbance. Similarly, the RMSE of the knee joint increases from $0.084^{\circ}$ to $0.789^{\circ}$ as the magnitude of the disturbance increases. The results show that the controller can maintain stable trajectory tracking even in the presence of large external disturbances.

\begin{table}[!ht]
\centering
\caption{Tracking performance (RMSE) of the proposed control design under external joint torque disturbances.}
\label{tab:robustness}
\small
\begin{tabular}{@{}cccc@{}}
\toprule
\textbf{Disturbance (Nm)} & \textbf{Hip RMSE (deg)} & \textbf{Knee RMSE (deg)} & \textbf{Ankle RMSE (deg)} \\
\midrule
0  & 0.124 & 0.084 & 0.047 \\
3  & 0.143 & 0.166 & 0.079 \\
6  & 0.172 & 0.312 & NA \\
9  & 0.218 & 0.524 & NA \\
12 & 0.285 & 0.789 & NA \\
\bottomrule
\end{tabular}
\end{table}

The ankle joint was evaluated up to 3~Nm, where RMSE increased from $0.047^{\circ}$ to $0.079^{\circ}$, remaining within acceptable bounds. Higher disturbance levels were not applicable at the ankle because the ankle actuator had a lower torque capacity than the hip and knee joints. Overall, the proposed controller demonstrates satisfactory robustness to external torque disturbances, with the online RBF weight adaptation and robust sub-controller term providing effective disturbance rejection at all three joints within their respective operating ranges.

\section{Conclusion}
\label{sec:conclusion}

This paper presented a model-free adaptive backstepping controller with a Gaussian RBF neural estimator and a TD3 supervisory gain scheduler for STS trajectory tracking of a bilateral lower-limb exoskeleton. The proposed controller was evaluated through co-simulation in MATLAB/Simulink and Simscape Multibody against PID, LQR, SMC, and MFAC benchmark controllers. Simulation results demonstrated that the proposed controller achieved superior trajectory-tracking performance across the hip, knee, and ankle joints, while maintaining smooth control actions and robustness against external disturbances. The key conclusions drawn from this study are as follows:

\begin{itemize}[leftmargin=*, itemsep=2pt]
\item The proposed RBF-MFAC backstepping controller achieves the lowest average RMSE of $0.078^{\circ}$ for all joints, with improvements of 60.2\%, 48.7\%, 42.6\%, and 54.4\% compared to PID, LQR, SMC, and MFAC, respectively, which demonstrates the effectiveness of online disturbance estimation over fixed-model and fixed-gain approaches.
\item The TD3-based supervisory gain scheduler enhanced the tracking performance of the baseline RBF-MFAC controller, yielding RMSE reductions of approximately 35\%, 33\%, and 79\% for the hip, knee, and ankle joints, respectively.
\item The proposed controller demonstrates the lowest actuator effort among all evaluated strategies, with hip peak torque of 28~Nm compared to 145~Nm for PID and 171~Nm for MFAC, showing that online disturbance compensation reduces the reliance on large gain-driven torque corrections.
\item Robustness analysis shows that the proposed controller maintains stable tracking performance under external torque disturbances up to 12~Nm at the hip and knee joints, with RMSE remaining below $0.3^{\circ}$ and $0.8^{\circ}$, respectively, due to the robust sub-controller and online RBF weight adaptation.
\end{itemize}

While the presented results validate the proposed control design in a high fidelity co-simulation environment, further investigation is needed to assess its practical applicability under real-world operating conditions. Future work will focus on validation using real human-subject motion capture data, asymmetric STS motion scenarios, real prototype stability analysis, hardware implementation in a physical exoskeleton prototype, and integration of physiological signals such as electromyography (EMG) for user-specific adaptive assistance during sit-to-stand movements.


\section*{Acknowledgments}

This work was supported by the Science and Engineering Research Board, Government of India, under Grant No.~CRG/2022/005578, and the Seed Money Grant (Project No.~DoRDC 763) from Thapar Institute of Engineering and Technology, Patiala, India.

\section*{Declaration of competing interest}

The authors declare that they have no known competing financial interests or personal relationships that could have appeared to influence the work reported in this paper.

\bibliography{ref}

@techreport{UN2017,
  author      = {{United Nations Department of Economic and Social Affairs Population Division}},
  title       = {World Population Prospects: The 2017 Revision},
  institution = {United Nations},
  address     = {New York, NY},
  year        = {2017}
}

@article{Shukla2020,
  author  = {Shukla, B. and Bassement, J. and Vijay, V. and Yadav, S. and Hewson, D.},
  title   = {Instrumented Analysis of the Sit-to-Stand Movement for Geriatric Screening: A Systematic Review},
  journal = {Bioengineering},
  volume  = {7},
  number  = {4},
  pages   = {139},
  year    = {2020}
}

@article{Wang2022,
  author  = {Wang, X. and Huang, Y. and Chen, Y. and Yang, T. and Su, W. and Chen, X. and Yan, F. and Han, L. and Ma, Y.},
  title   = {The Relationship between Body Mass Index and Stroke: A Systematic Review and Meta-Analysis},
  journal = {Journal of Neurology},
  volume  = {269},
  number  = {12},
  pages   = {6279--6289},
  year    = {2022}
}

@article{Feigin2022,
  author  = {Feigin, V. L. and Brainin, M. and Norrving, B. and Martins, S. and Sacco, R. L. and Hacke, W. and Fisher, M. and Pandian, J. and Lindsay, P.},
  title   = {World Stroke Organization (WSO): Global Stroke Fact Sheet 2022},
  journal = {International Journal of Stroke},
  volume  = {17},
  number  = {1},
  pages   = {18--29},
  year    = {2022}
}

@article{Young2016,
  author  = {Young, A. J. and Ferris, D. P.},
  title   = {State of the Art and Future Directions for Lower Limb Robotic Exoskeletons},
  journal = {IEEE Transactions on Neural Systems and Rehabilitation Engineering},
  volume  = {25},
  number  = {2},
  pages   = {171--182},
  year    = {2016}
}

@article{Dall2009,
  author  = {Dall, P. M. and Kerr, A.},
  title   = {Frequency of the Sit-to-Stand Task: An Observational Study of Free-Living Adults},
  journal = {Applied Ergonomics},
  volume  = {41},
  number  = {1},
  pages   = {58--61},
  year    = {2009}
}

@article{Alcazar2018,
  author  = {Alcazar, J. and Losa-Reyna, J. and Rodriguez-Lopez, C. and Alfaro-Acha, A. and Rodriguez-Ma{\~n}as, L. and Ara, I. and Garc{\'i}a-Garc{\'i}a, F. J. and Alegre, L. M.},
  title   = {The Sit-to-Stand Muscle Power Test: An Easy, Inexpensive, and Portable Procedure to Assess Muscle Power in Older People},
  journal = {Experimental Gerontology},
  volume  = {112},
  pages   = {38--43},
  year    = {2018}
}

@article{Pransky2014,
  author  = {Pransky, J.},
  title   = {The Pransky Interview: Russ Angold, Co-Founder and President of Ekso Labs},
  journal = {Industrial Robot},
  volume  = {41},
  number  = {4},
  pages   = {329--334},
  year    = {2014}
}

@article{Banala2010,
  author  = {Banala, S. K. and Agrawal, S. K. and Kim, S. H. and Scholz, J. P.},
  title   = {Novel Gait Adaptation and Neuromotor Training Results Using an Active Leg Exoskeleton},
  journal = {IEEE/ASME Transactions on Mechatronics},
  volume  = {15},
  number  = {2},
  pages   = {216--225},
  year    = {2010}
}

@inproceedings{Bernhardt2005,
  author    = {Bernhardt, M. and Frey, M. and Colombo, G. and Riener, R.},
  title     = {Hybrid Force-Position Control Yields Cooperative Behaviour of the Rehabilitation Robot {LOKOMAT}},
  booktitle = {Proceedings of the IEEE International Conference on Rehabilitation Robotics},
  pages     = {536--539},
  year      = {2005},
  doi       = {10.1109/ICORR.2005.1501159}
}

@article{Li2017,
  author  = {Li, Z. and Ma, W. and Yin, Z. and Guo, H.},
  title   = {Tracking Control of Time-Varying Knee Exoskeleton Disturbed by Interaction Torque},
  journal = {ISA Transactions},
  volume  = {71},
  pages   = {458--466},
  year    = {2017}
}

@article{Aliman2017,
  author  = {Aliman, N. and Ramli, R. and Haris, S. M.},
  title   = {Design and Development of Lower Limb Exoskeletons: A Survey},
  journal = {Robotics and Autonomous Systems},
  volume  = {95},
  pages   = {102--116},
  year    = {2017}
}

@article{Shepherd2017,
  author  = {Shepherd, M. K. and Rouse, E. J.},
  title   = {Design and Validation of a Torque-Controllable Knee Exoskeleton for Sit-to-Stand Assistance},
  journal = {IEEE/ASME Transactions on Mechatronics},
  volume  = {22},
  number  = {4},
  pages   = {1695--1704},
  year    = {2017}
}

@article{Vantilt2019,
  author  = {Vantilt, J. and Tanghe, K. and Afschrift, M. and Bruijnes, A. K. B. D. and Junius, K. and Geeroms, J. and Aertbeli{\"e}n, E. and De Groote, F. and Lefeber, D. and Jonkers, I. and De Schutter, J.},
  title   = {Model-Based Control for Exoskeletons with Series Elastic Actuators Evaluated on Sit-to-Stand Movements},
  journal = {Journal of NeuroEngineering and Rehabilitation},
  volume  = {16},
  number  = {1},
  pages   = {65},
  year    = {2019}
}

@article{Huo2021,
  author  = {Huo, W. and Moon, H. and Alouane, M. A. and Bonnet, V. and Huang, J. and Amirat, Y. and Vaidyanathan, R. and Mohammed, S.},
  title   = {Impedance Modulation Control of a Lower-Limb Exoskeleton to Assist Sit-to-Stand Movements},
  journal = {IEEE Transactions on Robotics},
  volume  = {38},
  number  = {2},
  pages   = {1230--1249},
  year    = {2021}
}

@article{Roelker2022,
  author  = {Roelker, S. A. and Schmitt, L. C. and Chaudhari, A. M. W. and Siston, R. A.},
  title   = {Discover Your Potential: The Influence of Kinematics on a Muscle's Ability to Contribute to the Sit-to-Stand Transfer},
  journal = {PLoS One},
  volume  = {17},
  number  = {3},
  pages   = {e0264080},
  year    = {2022}
}

@article{Fernandez2026,
  author  = {Fernandez-Montoya, M. and Erickson, E. J. and Gallego, J. A. and Aguirre, M. E.},
  title   = {Development and Verification of a Model to Predict the Adjustable Force Transmission Capabilities of a Passive, Lower Back Exoskeleton},
  journal = {ASME Journal of Mechanisms and Robotics},
  volume  = {18},
  number  = {5},
  pages   = {1--19},
  year    = {2026}
}

@article{Cheng2026,
  author  = {Cheng, G. and Huang, Y. and Zhang, X.},
  title   = {Dynamic Parameter Identification Using Triple-Loop Iteration for a Variable Stiffness Exoskeleton in Sit-to-Stand Assistance},
  journal = {Nonlinear Dynamics},
  volume  = {114},
  number  = {7},
  year    = {2026}
}

@article{Hernandez2020,
  author  = {Hern{\'a}ndez, J. H. and Cruz, S. S. and L{\'o}pez-Guti{\'e}rrez, R. and Gonz{\'a}lez-Mendoza, A. and Lozano, R.},
  title   = {Robust Nonsingular Fast Terminal Sliding-Mode Control for Sit-to-Stand Task Using a Mobile Lower Limb Exoskeleton},
  journal = {Control Engineering Practice},
  volume  = {101},
  pages   = {104496},
  year    = {2020}
}

@article{Narayan2024,
  author  = {Narayan, J. and Abbas, M. and Dwivedy, S. K.},
  title   = {Adaptive Backstepping Sliding Mode Subject-Cooperative Control for a Pediatric Lower-Limb Exoskeleton Robot},
  journal = {Transactions of the Institute of Measurement and Control},
  volume  = {47},
  number  = {2},
  pages   = {352--368},
  year    = {2024}
}

@article{Sharma2021,
  author  = {Sharma, R. and Gaur, P. and Bhatt, S. and Joshi, D.},
  title   = {Optimal Fuzzy Logic-Based Control Strategy for Lower Limb Rehabilitation Exoskeleton},
  journal = {Applied Soft Computing},
  volume  = {105},
  pages   = {107226},
  year    = {2021}
}

@article{Yang2020,
  author  = {Yang, S. and Han, J. and Xia, L. and Chen, Y.-H.},
  title   = {An Optimal Fuzzy-Theoretic Setting of Adaptive Robust Control Design for a Lower Limb Exoskeleton Robot System},
  journal = {Mechanical Systems and Signal Processing},
  volume  = {141},
  pages   = {106706},
  year    = {2020}
}

@article{Liu2019,
  author  = {Liu, X. and Zhou, Z. and Mai, J. and Wang, Q.},
  title   = {Real-Time Mode Recognition Based Assistive Torque Control of Bionic Knee Exoskeleton for Sit-to-Stand and Stand-to-Sit Transitions},
  journal = {Robotics and Autonomous Systems},
  volume  = {119},
  pages   = {209--220},
  year    = {2019}
}

@article{Fleischer2008,
  author  = {Fleischer, C. and Hommel, G.},
  title   = {A Human-Exoskeleton Interface Utilizing Electromyography},
  journal = {IEEE Transactions on Robotics},
  volume  = {24},
  number  = {4},
  pages   = {872--882},
  year    = {2008}
}

@article{Yu2026,
  author  = {Yu, S. and Fu, R. and Ye, C. and Li, H.},
  title   = {Adaptive Patient-Cooperative Control of an Omnidirectional Mobile Exoskeleton Robot for Lower Limb Rehabilitation},
  journal = {ASME Journal of Mechanisms and Robotics},
  volume  = {18},
  number  = {5},
  pages   = {1--31},
  year    = {2026}
}

@article{Fliess2013,
  author  = {Fliess, M. and Join, C.},
  title   = {Model-Free Control},
  journal = {International Journal of Control},
  volume  = {86},
  number  = {12},
  pages   = {2228--2252},
  year    = {2013}
}

@article{Amara2025,
  author  = {Amara, Y. and Kenas, F.},
  title   = {Adaptive Model-Free Control of Lower Limb Exoskeletons Using Neural Estimation and Swarm-Based Optimization},
  journal = {Journal of the Brazilian Society of Mechanical Sciences and Engineering},
  volume  = {47},
  number  = {12},
  year    = {2025}
}

@article{Khan2019,
  author  = {Khan, S. G. and Tufail, M. and Shah, S. H. and Ullah, I.},
  title   = {Reinforcement Learning Based Compliance Control of a Robotic Walk Assist Device},
  journal = {Advanced Robotics},
  volume  = {33},
  number  = {24},
  pages   = {1281--1292},
  year    = {2019}
}

@article{Delp2007,
  author  = {Delp, S. L. and Anderson, F. C. and Arnold, A. S. and Loan, P. and Habib, A. and John, C. T. and Guendelman, E. and Thelen, D. G.},
  title   = {{OpenSim}: Open-Source Software to Create and Analyze Dynamic Simulations of Movement},
  journal = {IEEE Transactions on Biomedical Engineering},
  volume  = {54},
  number  = {11},
  pages   = {1940--1950},
  year    = {2007}
}

@article{Kumbhar2026,
  author  = {Kumbhar, R. and Singh, R. and Gadade, A. M. and Singla, A. and Hussain, I.},
  title   = {A Novel Hybrid {PID-LQR} Controller for Sit-To-Stand Assistance Using a {CAD}-Integrated Simscape Multibody Lower Limb Exoskeleton},
  journal = {arXiv preprint},
  year    = {2026}
}

@article{Narayan2023,
  author  = {Narayan, J. and Abbas, M. and Patel, B. and Dwivedy, S. K.},
  title   = {Adaptive {RBF} Neural Network-Computed Torque Control for a Pediatric Gait Exoskeleton System: An Experimental Study},
  journal = {Intelligent Service Robotics},
  volume  = {16},
  number  = {5},
  pages   = {549--564},
  year    = {2023}
}

@article{Park1991,
  author  = {Park, J. and Sandberg, I. W.},
  title   = {Universal Approximation Using Radial-Basis-Function Networks},
  journal = {Neural Computation},
  volume  = {3},
  number  = {2},
  pages   = {246--257},
  year    = {1991}
}

@article{Moran2026,
  author  = {Moran-Armenta, M. and Aguilar-Ibanez, C. and Moreno-Valenzuela, J.},
  title   = {Dual Neural Networks and {PID} Framework for Adaptive Trajectory Tracking Control of Manipulators: A Convergence-Centric Approach},
  journal = {Cybernetics and Systems},
  pages   = {1--31},
  year    = {2026}
}

@inproceedings{Fujimoto2018,
  author    = {Fujimoto, S. and Hoof, H. and Meger, D.},
  title     = {Addressing Function Approximation Error in Actor-Critic Methods},
  booktitle = {Proceedings of the 35th International Conference on Machine Learning (ICML 2018)},
  address   = {Stockholm, Sweden},
  pages     = {1587--1596},
  year      = {2018}
}

\end{document}